\documentclass{article}

\usepackage{microtype}
\usepackage{graphicx}
\usepackage{booktabs} 
\usepackage{multirow}
\usepackage{multicol}
\usepackage{tablefootnote}
\usepackage{xcolor}
\usepackage{fix-cm} 
\DeclareMathSizes{10}{10}{6}{4}

\usepackage{hyperref}

\usepackage[accepted]{icml2023}

\usepackage{amsmath}
\usepackage{amssymb}
\usepackage{mathtools}
\usepackage{amsthm}
\usepackage{caption}
\usepackage{subcaption}
\usepackage{booktabs}
\usepackage[flushleft]{threeparttable}
\usepackage{MnSymbol}
\usepackage{lipsum}

\usepackage{enumitem}

\usepackage[capitalize,noabbrev]{cleveref}

\theoremstyle{plain}
\newtheorem{theorem}{Theorem}

\newtheorem{lemma}{Lemma}
\newtheorem{corollary}{Corollary}
\theoremstyle{definition}
\newtheorem{definition}{Definition}
\newtheorem{assumption}{Assumption}
\theoremstyle{remark}
\newtheorem{remark}{Remark}

\usepackage[textsize=tiny]{todonotes}

\usepackage{amsmath}
\usepackage{amssymb}
\usepackage{ulem}
\usepackage{bm}
\usepackage[flushleft]{threeparttable}

\newcommand\bolden[1]{{\boldmath\bfseries#1}}
\newcommand{\smallsection}[1]{{\vspace{0.02in} \noindent {\bolden{\uline{\smash{#1}}}}}}

\definecolor{lucky}{RGB}{120, 130, 150}

\newcommand\green[1]{\textcolor{green}{#1}}
\newcommand\yellow[1]{\textcolor{yellow}{#1}}

\newcommand{\bolita}[1]{\textbf{\textit{#1}}}

\usepackage{graphicx}
\usepackage{xspace}
\usepackage{amsthm}
\usepackage{amsfonts}

\def\mydefbb#1{\expandafter\def\csname bb#1\endcsname{\ensuremath{\mathbb{#1}}}}
\def\mydefallbb#1{\ifx#1\mydefallbb\else\mydefbb#1\expandafter\mydefallbb\fi}
\mydefallbb ABCDEFGHIJKLMNOPQRSTUVWXYZ\mydefallbb

\def\mydefcal#1{\expandafter\def\csname cal#1\endcsname{\ensuremath{\mathcal{#1}}}}
\def\mydefallcal#1{\ifx#1\mydefallcal\else\mydefcal#1\expandafter\mydefallcal\fi}
\mydefallcal ABCDEFGHIJKLMNOPQRSTUVWXYZ\mydefallcal

\newcommand{\norm}[1]{\left\lVert#1\right\rVert}
\newcommand{\abs}[1]{\left|#1\right|}
\newcommand{\set}[1]{\{#1\}}

\newcommand{\bus}[1]{\textbf{\uline{\smash{#1}}}}
\newcommand{\ours}{AERO-GNN\xspace}

\icmltitlerunning{Towards Deep Attention in Graph Neural Networks}

\begin{document}

\twocolumn[
\icmltitle{Towards Deep Attention in Graph Neural Networks: \\ Problems and Remedies}



\icmlsetsymbol{equal}{*}

\begin{icmlauthorlist}
\icmlauthor{Soo Yong Lee}{KAIST_AI}
\icmlauthor{Fanchen Bu}{KAIST_EE}
\icmlauthor{Jaemin Yoo}{CMU}
\icmlauthor{Kijung Shin}{KAIST_AI,KAIST_EE}
\end{icmlauthorlist}

\icmlaffiliation{KAIST_AI}{Kim Jaechul Graduate School of Artificial Intelligence, KAIST, Daejeon, Republic of Korea}
\icmlaffiliation{KAIST_EE}{School of Electrical Engineering, KAIST, Daejeon, Republic of Korea}
\icmlaffiliation{CMU}{Heinz College of Information Systems and Public Policy, Carnegie Mellon University, Pittsburgh, PA, USA}

\icmlcorrespondingauthor{Kijung Shin}{kijungs@kaist.ac.kr}

\icmlkeywords{Graph Neural Network, Graph Attention, Deep Graph Learning, Heterophily}

\vskip 0.3in
]



\printAffiliationsAndNotice{}  

\begin{abstract}

Graph neural networks (GNNs) learn the representation of graph-structured data, and their expressiveness can be further enhanced by inferring node relations for propagation. 
Attention-based GNNs infer neighbor importance to manipulate the weight of its propagation. 
Despite their popularity, the discussion on deep graph attention and its unique challenges has been limited. 
In this work, we investigate some problematic phenomena related to deep graph attention, including vulnerability to over-smoothed features and smooth cumulative attention.
Through theoretical and empirical analyses, we show that various attention-based GNNs suffer from these problems. 
Motivated by our findings, we propose \ours, a novel GNN architecture designed for deep graph attention. 
\ours provably mitigates the proposed problems of deep graph attention, which is further empirically demonstrated with (\textbf{a}) its adaptive and less smooth attention functions and (\textbf{b}) higher performance at deep layers (up to 64).
On 9 out of 12 node classification benchmarks, \ours outperforms the baseline GNNs, highlighting the advantages of deep graph attention.
Our code is available at \url{https://github.com/syleeheal/AERO-GNN}.
\end{abstract}

    \section{Introduction}
    \label{sec:intro}
    Graph neural networks (GNNs) are a class of neural networks for representation learning on graph-structured data. 
Recently, GNNs have been successfully applied to a wide range of graph-related tasks, including social influence prediction~\cite{gnn_social_influence}, traffic forecast~\cite{gnn_traffic}, physical system modeling~\cite{gnn-physics}, product recommendation~\cite{gnn_recommend}, and drug discovery~\cite{drug-discovery}.

GNNs widely adopt message-passing frameworks, composed of two main pillars: feature transformation and propagation (a.k.a. neighborhood aggregation)~\cite{sgc, messagepassingNN}. 
Feature transformation updates node features from previous layers' features.
Propagation involves each node passing its own features to its neighbors,
and for each node, the passed-down neighbor features are aggregated to update its own node features.
In this framework, a propagation layer determines the propagation weight for each adjacent node pair, and each additional layer allows nodes to propagate to one more hop of neighbors.

Attention-based GNNs aim to learn to propagate by inferring the relational importance between node pairs.
Graph attention infers relational importance for node pairs.
Among many, GAT and its variants~\cite{gat, cgat, gatv2, gt, supergat, fagcn, magna, dmp, cat} learn edge attention.
The \textit{edge-attention} models infer the importance, or weight, of each neighbor w.r.t. each source node by applying an attention mechanism between adjacent node pairs.
During propagation, each source node aggregates features from its neighbors based on the inferred importance. 

Another class of GNNs, which we call \textit{hop-attention} (or \textit{layer-attention}) models, learns the relative importance of each hop~\cite{dagnn, gprgnn, evaluatedeepGNN, asgc}.
Hop-attention models apply attention coefficients at every propagation layer to express the relative importance of a given hop in determining the final node features. 
Thereby, hop-attention models learn which hops, among multi-hop neighbors, each source node should attend to during propagation.
Intuitively, edge-attention models learn importance \textit{within} each hop, and hop-attention models learn importance \textit{of} each hop.

Can the existing attention-based GNNs learn expressive attention over deep layers? 
Prior research does not provide a clear answer.
While many studies report the over-smoothing problem of \textit{node features} at deep layers~\cite{oversmoothing, oversmoothing_natural, gcnii, dagnn},
they do not address how it may relate to \textit{graph attention}.
Some works focus on designing expressive graph attention layers~\cite{gatv2, gt, supergat, fagcn, dmp, cat}, but their scope is limited to shallow model depth.
Even for attention-based GNNs that generalize their performance over deep layers~\cite{dagnn, magna, gprgnn, evaluatedeepGNN}, they 
do not explicitly discuss properties related to deep attention. 
Very recently, \citet{deep-graph-transformer} explored the relationship between model depth and attention function for graph transformers. 
Their analysis, however, was confined to the relationship between model depth and transformer-style attention to substructures.
In short, the theoretical underpinnings to understand deep graph attention are underexplored.

In this work, we investigate some problematic phenomena related to deep graph attention. 
Through both theoretical and empirical analyses,
we show that the \textit{vulnerability to over-smoothed features} and \textit{smooth cumulative attention} limit the graph attention from becoming expressive over deep layers.
Specifically, several representative attention-based GNNs, including GATv2~\cite{gatv2}, FAGCN~\cite{fagcn}, GPRGNN~\cite{gprgnn}, and DAGNN~\cite{dagnn}, suffer from these problems.

Motivated by our analyses, we propose a novel graph attention architecture, 
\bus{A}ttentive d\bus{E}ep p\bus{RO}pagation-GNN (\ours).
We theoretically demonstrate that \ours can mitigate the stated problems,
which is further elaborated empirically by \ours's (\textbf{a}) adaptive and less smooth attention functions and (\textbf{b}) higher performance at deep layers (up to 64).
On 9 out of 12 node classification benchmarks, including both homophilic and heterophilic graphs, \ours outperforms all the baselines, highlighting the advantages of deep graph attention.


In summary, our central contributions are two-fold:
\begin{enumerate}
    \item \textbf{Theoretical Findings.} We formulate two theoretical limitations of deep graph attention. \ours provably mitigates the problems, whereas the representative attention-based GNNs inevitably suffer from them.
    \item \textbf{Empirical Findings.} \ours shows superior performance in node classification benchmarks. Also, compared to the representative attention-based GNNs, \ours learns more adaptive and less smooth attention functions at deep layers. 
\end{enumerate}

    \section{Preliminaries}
    \label{sec:preliminaries_and_related_works}
    \smallsection{Graphs.} 
Let $G = (V, E)$ be a graph with node set $V$ and  edge set $E \subseteq \binom{V}{2}$.\footnote{We assume undirected, unweighted graphs, but our theoretical results can be easily extended to directed and/or weighted graphs.}
Let $n = |V|$ and $m = |E|$ denote the number of nodes and edges, resp.
WLOG, we assume $V = [n] = \set{1, 2, \ldots, n}$.
Let $A = A(G) \in \set{0, 1}^{n \times n}$ denote the adjacency matrix of $G$, where $A_{ij} = 1$ iff $(i, j) \in E$, and we use $D = \operatorname{diag}(d_1, d_2, \ldots, d_n)$ to denote the degree matrix of $G$, where $d_i$ is the degree of node $i$.
Let $X \in \bbR^{n \times d_{x}}$ denote the initial node feature matrix, where each node $i \in V$ has node feature $X_i$ of dimension $d_{x}$. 

\smallsection{Message-Passing GNNs.}
Feature transformation and propagation are the two main building blocks of message-passing GNNs~\cite{sgc, messagepassingNN}.
Feature transformation updates the features of each node based on the node's features of previous layers. 
A feature transformation layer can be expressed as:
$H^{(k)} = \sigma(H^{(k-1)}W^{(k)}), \forall 1 \leq k \leq k_{max}$,\footnote{{Usually, $H^{(0)}$ is a function on the initial features, i.e., $X$.}}
where 
$k$ denotes the index of each layer with $k_{max}$ being the total number of layers,
$H^{(k)} \in \bbR^{n \times d_H}$ is the \textit{hidden node feature matrix} at layer $k$,
$W^{(k)}$ is the weight matrix at layer $k$,
and $\sigma$ is an activation function.

On the other hand, a propagation layer passes node features to its neighbors, and each node's features are updated as an aggregation of the features. 
Specifically, a propagation layer can be expressed as follows:
$H^{(k)} = \tilde{A}H^{(k-1)}, \forall 1 \leq k \leq k_{max}$,
where $\Tilde{A} = (D+I)^{-1/2} (A+I)(D+I)^{-1/2}$ is the symmetrically normalized adjacency matrix with self-loops.

Many GNNs fuse the two operations in one layer, such that $H^{(k)}=\sigma(\tilde{A} H^{(k-1)}W^{(k)})$~\cite{gcn, gat, gcnii}. Some others use only propagation layers to update $H^{(k)}$,
after obtaining $H^{(0)}$ with feature transformation~\cite{appnp, dagnn, gprgnn}. Note that $\tilde{A}$ determines the magnitude in which each node's features propagate to its neighbors, and the number of propagation layers determines the number of hops to propagate to.

\begin{table*}[t]
\centering
\caption{Attention Functions of Graph Attention Models} \label{tab:attention_functions}
    \scalebox{0.90}{
    \renewcommand{\arraystretch}{1.0}
\begin{threeparttable}

    \centering
    

        \begin{tabular}{c l l l}
            \toprule
            
            \multicolumn{1}{c}{Model} 
            & \multicolumn{1}{c}{Pre-normalized Edge Attention} 
            & \multicolumn{1}{c}{Normalized Edge Attention} 
            & \multicolumn{1}{c}{Hop Attention} 
            \\
            
            \midrule
            \midrule
            
            \textbf{GATv2}  & $ \check{\alpha}^{(k)}_{ij} = \exp((W^{(k)}_{edge})^{\top} \sigma (H_i^{(k-1)} \Vert H_j^{(k-1)})) $ 
                            & $ \alpha^{(k)}_{ij} = {\check{\alpha}^{(k)}_{ij}} / {\sum_{j' \in N(i)} \check{\alpha}^{(k)}_{ij'}} $
                            & $ \gamma^{(k)}_i = 1 $  
                        \\
            
            \textbf{FAGCN}  & $ \check{\alpha}^{(k)}_{ij} = \tanh((W^{(k)}_{edge})^{\top} (Z_i^{(k-1)} \Vert Z_j^{(k-1)})) $ 
                            & $ \alpha^{(k)}_{ij} = {\check{\alpha}^{(k)}_{ij}} / {\sqrt{d_i d_j}}$
                            & $ \gamma^{(k)}_i = c_{\gamma}, \forall{k,i}$ (*)
                        \\
            
            \textbf{GPRGNN} & $ \check{\alpha}^{(k)}_{ij} = 1 $ 
                            & $ \alpha^{(k)}_{ij} = {1} / {\sqrt{d_i d_j}}$
                            & $ \gamma^{(k)}_i = \gamma^{(k)}_j = c_{k; \gamma}, \forall k$ (**)
                        \\

            
            \textbf{DAGNN}  & $ \check{\alpha}^{(k)}_{ij} = 1 $ 
                            & $ \alpha^{(k)}_{ij} = {1} / {\sqrt{d_i d_j}}$
                            & $ \gamma^{(k)}_i = \operatorname{sigmoid}( {W_{hop}}^{\top} H_i^{(k)} ) $ (***)
                        \\

            

            \bottomrule
        \end{tabular}        
        \begin{tablenotes}
        \small
        \item (*) for FAGCN, $c_\gamma$ is a hyperparameter fixed before training and is identical for all $k$ and $i$
        \item (**) for GPRGNN, $c_{k; \gamma}$ is a learnable variable that can be different for different $k$, but is identical for all $i$ given any fixed $k$
        \item (***) for DAGNN, the same $W_{hop}$ is used for each layer        
    \end{tablenotes}                
\end{threeparttable}
}
\end{table*}

\begin{table*}[t]
\vspace{-2mm}
\caption{Propagation of Graph Attention Models} \label{tab: T}
    \centering
    \scalebox{0.90}{
    \renewcommand{\arraystretch}{1.0}
    \begin{threeparttable}
        \centering
        \begin{tabular}{c l l}
            \toprule
            
            \multicolumn{1}{c}{Model} 
            & \multicolumn{1}{c}{$k$-Layer Propagation} 
            & \multicolumn{1}{c}{Rephrased Propagation w.r.t $T^{(k)}$'s} 
            \\
            
            \midrule
            \midrule
            \textbf{GATv2}
                        & $H^{(k)} = (\prod_{\ell = k}^1 \mathcal{A}^{(\ell)}) X (\prod_{\ell' = 1}^k W^{(\ell')})$ 
                        & $H^{(k)} = T^{(k)} X \prod_{\ell = 1}^k W^{(\ell)} $
                        \\
            
            \textbf{FAGCN}  
            & $Z^{(k)} = (\Gamma + \Gamma \mathcal{A}^{(k)} + \Gamma \mathcal{A}^{(k)}\mathcal{A}^{(k-1)} +\cdots+ \Gamma \prod_{\ell=k}^1 \mathcal{A}^{(\ell)} )H^{(0)}$ (*)
            &  $Z^{(k)} = \sum_{\ell = 0}^k T^{(\ell)} H^{(0)} $\\
            
            \textbf{GPRGNN/DAGNN} 
            & $Z^{(k)} = ( \Gamma^{(0)} + \Gamma^{(1)} \mathcal{A} + \Gamma^{(2)} \mathcal{A}^2 +\cdots+ \Gamma^{(k)} \mathcal{A}^k )H^{(0)} $ (**)
            &  $Z^{(k)} = \sum_{\ell = 0}^k T^{(\ell)} H^{(0)}$\\

            
            

            \bottomrule
        \end{tabular}
        \begin{tablenotes}
        \small
        \item (*) recall that for FAGCN, $\Gamma^{(k)}$ is identical for all $k$, and we use a simplified notation $\Gamma$ to denote all $\Gamma^{(k)}$'s
        \item (**) recall that for GPRGNN and DAGNN, $\calA^{(k)}$ is identical for all $k$, and we use $\calA$ to denote all $\calA^{(k)}$'s (here, $\calA^{k}$ is the $k$-th power of $\calA$)
    \end{tablenotes}   
        \end{threeparttable}
        }
        \vspace{3mm}
\end{table*}

\smallsection{Edge Attention.} 
{Edge-attention GNNs (e.g., GAT and its variants) learn an edge-attention matrix $\mathcal{A}^{(k)} = (\alpha^{(k)}_{ij}) \in \mathbb{R}^{n \times n}$ at each propagation layer (i.e., hop) $k$,\footnote{Single-head attention is assumed for simplicity.}
where each edge attention coefficient $\alpha_{ij}^{(k)}$ can be seen as the importance of node $j$ w.r.t node $i$ at layer $k$.
At each propagation layer, the edge attention coefficients are used to weigh propagation between each adjacent node pair.}

\smallsection{Hop Attention.}
Hop-attention GNNs learn a hop-attention matrix $\Gamma^{(k)} = \operatorname{diag}(\gamma_1^{(k)}, \gamma_2^{(k)}, \ldots, \gamma_n^{(k)}) \in \bbR^{n \times n}$ at each propagation layer $k$.
With hop attention, different importance $\gamma^{(k)}_i$ can be assigned at different layers $k$ for every node $i$.
Typically, 
\begin{align} 
    Z^{(k)} &= \sum\nolimits_{\ell=0}^{k} \Gamma^{(\ell)}H^{(\ell)}, \forall 1 \leq k \leq k_{max}, \\
    &= \sum\nolimits_{\ell=0}^{k} \Gamma^{(\ell)} \Tilde{A}^{\ell} H^{(0)}, \forall 1 \leq k \leq k_{max}, 
\end{align}
\vspace{2mm}

where the sum of hidden feature matrix $H^{(\ell)}$'s up to layer $k$, each weighted by the hop attention matrix $\Gamma^{(\ell)}$, expresses the \textit{layer-aggregated} node feature matrix $Z^{(k)}$. Equation (2) highlights that the hop attention matrix $\Gamma^{(\ell)}$ can be seen as learning weights of $\ell$ hop neighbors expressed in $\Tilde{A}^\ell$ (A similar analysis can be found in ~\citet{equivalence}).

    \section{Theoretical Analysis on Deep Attention}
    \label{sec:theory}
    Can attention functions of the representative attention-based GNNs remain expressive over deeper layers?
Prior research has suggested possible reasons for the performance degradation of deep GNNs, including the over-smoothing of node features, over-squashing~\cite{bottleneck}, and over-correlation~\cite{overcorrelation}.
However, discussion dedicated to theoretical limitations of \textit{deep graph attention} has been little. In this section, we formulate two theoretical limitations of the representative attention-based GNNs concerning their ability to remain expressive over deeper layers.

\subsection{A Systematic Understanding of Graph Attention}
\label{subsec:new_understanding_of_attention}
{A systematic understanding of graph attention, integrating edge and hop attention, would allow us to discuss various attention-based GNNs and their theoretical limitations within the same framework.}

\smallsection{Cumulative Attention.}
We have observed that edge and hop attention achieve the same goal (i.e., inferring the relational importance between node pairs), but from two different perspectives. 
Consequently, they both learn weights in which each node's features propagate to its neighbors.
This observation motivates us to integrate edge and hop attention under the same umbrella.
To this end, we propose a concept of \textit{cumulative attention matrix}, denoted by $T^{(k)}$, which intuitively represents attention between all node pairs within $k$ hops (or equivalently, at layer $k$) that considers both edge and hop attentions.

Formally, given any $k_{max}$, for each $0 \leq k \leq k_{max}$,
\begin{equation}\label{eq:cumulative attention matrix}
    T^{(k)} =  \Gamma^{(k)} \prod\nolimits_{\ell = k}^1 \mathcal{A}^{(\ell)},
\end{equation}
where $T^{(0)} = \Gamma^{(0)}$.\footnote{For FAGCN, $T^{(k)} = \Gamma^{(k)} \prod\nolimits_{\ell = k_{max}}^{k_{max} - k + 1} \mathcal{A}^{(\ell)}$. However, in our theoretical analyses, it is equivalent to Eq.~\eqref{eq:cumulative attention matrix} when $k_{max}$ is fixed.}

\smallsection{Rephrased Propagation w.r.t $T^{(k)}$'s.}
Here, we discuss various representative attention-based GNNs (spec., GATv2, FAGCN, GPRGNN, and DAGNN).
We detail their attention functions in Table~\ref{tab:attention_functions}.
Those models are used throughout our theoretical analyses and empirical evaluation.

GATv2 and FAGCN learn edge attention $\alpha^{(k)}_{ij}$. GATv2 uses hidden features ($H_i \Vert H_j$), whereas FAGCN layer-aggregated features ($Z_i \Vert Z_j$), in computing each edge attention coefficient $\alpha^{(k)}_{ij}$. Notably, their $\alpha^{(k)}_{ij}$ have different bounds, such that GATv2's $\alpha^{(k)}_{ij} \in (0, 1)$ and FAGCN's $\alpha^{(k)}_{ij} \in (-1, 1)$. They do not learn hop attention coefficients, which can be expressed as a constant (1 for GATv2 and $c_{\gamma}$ for FAGCN).




On the other hand, GPRGNN and DAGNN learn hop attention $\gamma_i^{(k)}$. GPRGNN's hop attention is not explicitly bounded and, thus, can learn negative hop attention $\gamma_i^{(k)}$. While $\gamma_i^{(k)} \in (0, 1)$ for DAGNN, it learns \textit{node-adaptive} hop attention.
Neither GPRGNN nor DAGNN learns nontrivial edge attention, and their edge attention coefficients are equivalently degree-normalized constants,
i.e., $\alpha^{(k)}_{ij} = 1 / \sqrt{d_i d_j}, \forall i,j,k$.


The discussed GNNs consist of a representative and diverse set of attention functions, yet we can succinctly rephrase them w.r.t $T^{(k)}$ (see Table~\ref{tab: T}).\footnote{FAGCN and GATv2's propagation formulae are rewritten from their original form. Refer to Appendix~\ref{app:rewrite_msg_passing} for the details.}

\subsection{Theoretical Problems of Deep Graph Attention}
Two assumptions are used throughout our theoretical analyses, and proper normalization and preprocessing may always satisfy them in practice. 
\begin{assumption}\label{aspt:basic}
    The graph $G$ is connected and non-bipartite, and the initial node features in $X$ are pairwise distinct (i.e., $X_i \neq X_j, \forall i \neq j \in V$) and finite 
    (i.e., $\norm{X_i}_F < \infty, \forall i \in V$).
\end{assumption}

\begin{assumption}\label{aspt:parameter_bounded}
    There exists a global constant $C_{param} > 0$ such that,
    for each considered GNN model,
    each entry of each parameter matrix or vector (e.g., $W^{(k)}$'s)    
    and each entry of each intermediate variable (e.g., $H^{(k)}$'s and $Z^{(k)}$'s)
    is bounded in $[-C_{param}, C_{param}]$.
\end{assumption}

\smallsection{Problem 1: Vulnerability to Over-Smoothing.}
In the first problem, we establish a connection between attention functions\footnote{We see attention matrices as functions here.} and the node feature over-smoothing.
Specifically, we examine the \textit{vulnerability} and \textit{resistance} of attention functions to over-smoothed node features. 
Informally, if the pre-normalized edge attention coefficients $\check{\alpha}^{(k)}_{ij}$'s (resp., hop attention coefficients $\gamma^{(k)}_{i}$'s) are always identical for the node pairs (resp., nodes) with identical (over-smoothed) hidden features, the attention function is vulnerable to over-smoothing.
\footnote{We use such a definition using the \textit{equality} of node features for simplicity. See Appendix~\ref{app:def_v2os_relax} for a relaxed version of Definition~\ref{def:vul_to_over_smoothing}.}
For simplicity, we use $f_{att} = f_{att}(\theta)$ to denote $(\check{\calA}, \Gamma)$, where $\theta$ denotes GNN parameters.
\vspace{0.5mm}
\begin{definition}\label{def:vul_to_over_smoothing}
    Given $G = (V, E)$ and initial node features $X$ satisfying Assumption~\ref{aspt:basic},
    suppose that $H^{(k')}_{i} = H^{(k')}_{i'}$ and $H^{(k')}_{j} = H^{(k')}_{j'}$ holds (due to the over-smoothing of node features) for some $(i, j), (i', j') \in E$ and $k' \geq 1$.
    We say that $f_{att}$ is \bolita{vulnerable to over-smoothing} (V2OS),
    if $\forall \theta$, $\check{\alpha}^{(k'+1)}_{ij} = \check{\alpha}^{(k'+1)}_{i'j'}$ and $\gamma^{(k')}_{i} = \gamma^{(k')}_{i'}$;
    that $f_{att}$ is \bolita{weakly resistant to over-smoothing} (WR2OS), 
    if $\exists \theta$, $\check{\alpha}^{(k'+1)}_{i j} \neq \check{\alpha}^{(k'+1)}_{i' j'}$ or $\gamma^{(k')}_{i} \neq \gamma^{(k')}_{i'}$;
    and that $f_{att}$ is \bolita{strongly resistant to over-smoothing} (SR2OS), 
    if $\exists \theta$, $\check{\alpha}^{(k'+1)}_{i j} \neq \check{\alpha}^{(k'+1)}_{i' j'}$ and $\gamma^{(k')}_{i} \neq \gamma^{(k')}_{i'}$.
\end{definition}
\vspace{0.5mm}
\begin{remark}\label{rem:significance_of_R2OS} 
    Definition 1 is practically significant. 
    A GNN with WR2OS/SR2OS $f_{att}$ can possibly remain expressive, even when some node features in intermediate layers begin to over-smooth.
    An expressive $f_{att}$ can mitigate widely known problems of deep GNNs, including over-smoothing and over-squashing.
\end{remark}
\vspace{0.5mm}
\begin{theorem}\label{thm:vul_to_os}
    For GATv2, GPRGNN, and DAGNN, $f_{att}$ is V2OS;
    for FAGCN, $f_{att}$ is WR2OS (but not SR2OS).
\end{theorem}
\vspace{1mm}
\begin{proof}
    All the proofs are in Appendix~\ref{app:proofs}.
\end{proof}

\smallsection{Problem 2: Smooth Cumulative Attention.}
If Problem 1 does not exist, such that the attention coefficients are non-trivial for any model depth, can the attention functions remain expressive over deeper layers?
We argue not.
For the representative attention-based GNNs, we show that the cumulative attention matrices $T^{(k)}$'s become over-smoothed over increasing layers, such that different nodes have attention close to each other, up to a positive scaling factor.
\textit{This is critically contrary to the goal of attention.}
Formally, we define a smoothness score $S: \bbR^{n \times n} \to \bbR_{\geq 0}$ by
\begin{equation}\label{eq:smoothness_func}
    S(T) = {\sum\nolimits_{(i, j) \in \binom{[n]}{2}} \norm{\frac{T^{(k)}_i}{\norm{T^{(k)}_i}}_1 - \frac{T^{(k)}_j}{\norm{T^{(k)}_j}_1}}_1} / {\binom{n}{2}}.\footnote{If $T_i$ is a zero vector, we let $T_i / \norm{T_i}_1$ be a zero vector too. The definition is similar to the smoothness metric defined in~\citet{dagnn}.}
\end{equation}
$S(\cdot)$ is bounded,
and a smaller $S(T^{(k)})$ indicates that $T^{(k)}$ is smoother.
Specifically, when $S(T^{(k)}) = 0$, all the rows of $T^{(k)}$ become equivalent up to a positive scaling factor.
See Appendix~\ref{app:proofs} for more theoretical properties of $S(\cdot)$.

\begin{theorem}\label{thm:smooth_T}
    Given $G = (V, E)$ and $X$ with Assumptions~\ref{aspt:basic} and ~\ref{aspt:parameter_bounded} satisfied,
    for GATv2, GPRGNN, and DAGNN, $\lim_{k \to \infty} S(T^{(k)}) = 0$;
    for FAGCN, $\lim_{k \to \infty} T^{(k)}_{ij} = 0, \forall i, j$.
\end{theorem}
By Theorem~\ref{thm:smooth_T}, for the aforementioned attention-based GNNs, $S(T^{(k)})$ converges to 0 and loses expressiveness as $k$ goes to infinity. 
Importantly, in our proofs, we show that problem 2 occurs \textit{without assuming problem 1}.

Below, we provide some insights into the reason why the GNNs without \textit{node-adaptive hop attention} or \textit{negative attention} may suffer from the problem stated above.
\vspace{0.5mm}
\begin{definition}\label{def:attention path}
    Given $G = (V, E)$, and $i, j \in V$,
    the set of \bolita{attention path} between $i$ and $j$ at layer $k$, denoted as $\calP^{(k)}(i, j)$, is defined as
    $\set{(i = v_0, v_1, v_2, \ldots, v_{k-1}, v_k = j): (v_{l - 1}, v_l) \in E, \forall 1 \leq l \leq k}$, i.e., the set of all the paths of length $k$ from $i$ to $j$ in $G$.\footnote{Unlike \textit{simple} paths, repeated nodes are allowed.}
    The \bolita{degree of intersection} between two paths
    $\mathbf{p}_v = (v_0, v_1, v_2, \ldots, v_{k-1}, v_k)$ and
    $\mathbf{p}_u = (u_0, u_1, u_2, \ldots, u_{k-1}, u_k)$, denoted by $\operatorname{doi}(\mathbf{v}, \mathbf{u})$, is defined as 
    $|\set{t: 1 \leq t \leq k, v_{t-1} = u_{t-1}, v_t = u_t}|$.
\end{definition}
\vspace{0.5mm}
\begin{remark}\label{rem:intuition_att_paths}
    The intuition of Definition 2 is that we can decompose each entry $T^{(k)}_{ij}$ with respect to attention paths from $j$ to $i$.
    Specifically,
    $T^{(k)}_{ij} = \gamma^{(k)}_i \sum_{(j = v_0, v_1, \ldots, v_k = i) \in \calP^{(k)}(j, i)} \prod_{\ell = 1}^{k} {\alpha}^{(\ell)}_{v_{\ell-1}, v_\ell}, \forall i, j, k.$
\end{remark}
\vspace{0.5mm}

\begin{lemma}\label{lem:many_long_intersections}
    Given $G = (V, E)$ and initial node features satisfying Assumption~\ref{aspt:basic}, $i, j, x \in V$, and any $N_1, N_2 > 0$, there exists $K$ such that $|\set{(\mathbf{p}_i, \mathbf{p}_j): \mathbf{p}_i \in \calP^{(k)}(i, x), \mathbf{p}_j \in \calP^{(k)}(j, x), \operatorname{doi}(\mathbf{p}_i, \mathbf{p}_j) \geq N_1}| \geq N_2, \forall k \geq K$.
\end{lemma}

\begin{proof}
    All the proofs are in Appendix~\ref{app:proofs}.
\end{proof}
By Lemma~\ref{lem:many_long_intersections} and Remark~\ref{rem:intuition_att_paths}, 
the constituent terms of $T_{ix}$ and $T_{jx}$ increasingly intersect at deeper layers for any $i, j, x$.
In other words, bounds of $f_{att}$ and growing degree of intersection in $T^{(k)}$ together can cause problem 2, irrespective of $f_{att}$'s expressive power at each layer.
This partially explains why some GNNs without node-adaptive hop attention or negative attention end up with a smooth cumulative attention $T^{(k)}$. 
More details can be found in the proof of Theorem~\ref{thm:smooth_T} in Appendix~\ref{app:proofs}.

While we focused on GATv2, which is a representative variant of GAT, in our theoretical analysis, our analysis can be easily extended to other GAT variants, such as
SuperGAT~\cite{supergat} with a self-supervised loss term,
and CATs~\cite{cat} further using structural features.
See Appendix~\ref{app:proofs} for more discussions on those variants.
    
    \section{Proposed Method: \ours}
    \label{sec:method}
    In this section, we introduce the proposed model,
\bus{A}ttentive d\bus{E}ep p\bus{RO}pagation-GNN (\ours).
We present the model overview and discuss its attention functions.
Finally, we show how \ours provably addresses the theoretical limitations, with a discussion on model complexity (spec., number of parameters).

\subsection{Model Overview}\label{subsec:AEROGNN_overview}
The feature transformation and propagation of \ours consist of:
\begin{align}
    H^{(k)} & = 
\begin{cases}
   \operatorname{MLP}(X), & \text{if } k = 0, \\
    \calA^{(k)} H^{(k-1)},& \text{if }  1 \leq k \leq k_{max}, \\
\end{cases} \\
    Z^{(k)} &= \sum\nolimits_{\ell = 0}^{k} {\Gamma}^{(\ell)} H^{(\ell)}, \forall 1 \leq k \leq k_{max}, \label{eq: AERO-GNN_Z} \\
    Z^* &= \sigma (Z^{(k_{max})})W^*, \label{eq: AERO-GNN_Z*}
\end{align}
where $\operatorname{MLP}$ is a multi-layer perceptron for the feature transformation, $k_{max}$ is the total number of layers, $W^*$ is a learnable weight matrix, $Z^*$ is the final output node features, and $\sigma = \operatorname{ELU}$~\cite{elu} is the activation function.

\ours computes the edge attention matrix $\calA^{(k)}$ and the hop attention matrix ${\Gamma}^{(k)}$ with learnable parameters.
The propagation of \ours can also be written in terms of the cumulative attention matrices $T^{(k)}$'s 
in Eq.~\eqref{eq:cumulative attention matrix}, like the other attention-based GNNs
(see Table~\ref{tab: T}).
Specifically, 
\begin{align} 
    Z^{(k)} = \sum\nolimits_{\ell = 0}^k T^{(\ell)} H^{(0)}, \forall 1 \leq k \leq k_{max}.
\end{align}



\subsection{Using Layer-Aggregated Features}\label{subsec:AEROGNN_and_others}

We design \ours to use the layer-aggregated features $Z^{(k)}$ in computing both the edge attention $\calA^{(k)}$ and the hop attention ${\Gamma}^{(k)}$ at each layer $k$ to make it resistant to over-smoothing.
In Theorem~\ref{thm:vul_to_os}, FAGCN is the only model that is WR2OS (\textit{weakly resistant to over-smoothing}), since it uses $Z^{(k)}$'s for computing its edge attention (see Table~\ref{tab:attention_functions}).
Even if the node features become over-smoothed at deep layers, a GNN using layer-aggregated features $Z^{(k)}$ can adjust the edge (and hop) attention coefficients based on the cumulative information over multiple layers to allow attention functions to be non-trivial.

However, the magnitude of $Z^{(k)}$ may increase as $k$ increases, as shown in Eq.~\eqref{eq: AERO-GNN_Z}, which may cause instability.
We, thus, utilize weight-decay~\cite{gcnii} to re-scale $Z$.
Formally, the re-scaling is done by
\begin{equation*}
\begin{cases}
    \lambda_k &= \log(\frac{\lambda}{k} + 1 + \epsilon) \\
    \tilde{Z}^{(k)} &= \lambda_k Z^{(k)}
\end{cases}
\end{equation*}
where $\lambda > 0$ is a hyperparameter, and $\epsilon > 0$ is a small number (we use $\epsilon = 10^{-6}$) ensuring that $\lambda_k$ does not converge to $0$. We use $\tilde{Z}^{(k)}$ in the computation of both attention functions.

\subsection{Attention Functions}\label{sec:att_funcs}

\smallsection{Edge Attention.}
At every layer $1 \leq k \leq k_{max}$, we compute the pre-normalized edge attention $\check{\calA}^{(k)} = (\check{\alpha}^{(k)}_{ij})$ and the (normalized) edge attention $\calA^{(k)} = (\alpha^{(k)}_{ij})$ as follows:
\begin{equation*} 
\begin{cases} 
    \check{\alpha}^{(k)}_{ij} &= \operatorname{softplus}((W^{(k)}_{edge})^{\top} \sigma(\tilde{Z}^{(k-1)}_i \Vert \tilde{Z}^{(k-1)}_j)) \\
    \alpha^{(k)}_{ij} &= {\check{\alpha}^{(k)}_{ij}} / {\sqrt{\sum\nolimits_{j' \in N(i)} \check{\alpha}^{(k)}_{ij'} \sum\nolimits_{i' \in N(j)} \check{\alpha}^{(k)}_{ji'}}}
\end{cases}
\end{equation*}
where 
$W^{(k)}_{edge}$ is a learnable weight vector,
$\tilde{Z}^{(k-1)}_i$ is the $i$-th row of $\tilde{Z}^{(k-1)}$,
and $\calA^{(k)}$ is symmetrically normalized.

In \ours, we use the symmetric normalization, instead of the row-wise normalization used in GATv2, due to its theoretical and empirical superiority~\cite{wang2018attack,wang2021bag,he2020lightgcn}, especially w.r.t. training stability.
$\operatorname{Softplus}$~\cite{zheng2015improving} is used to positively map edge attention, with two primary advantages over two other mapping functions, $\operatorname{exp}$ and $\operatorname{tanh}$. Compared to $\operatorname{exp}$ used in GATv2, $\operatorname{softplus}$ has the higher computational stability ~\cite{kleshchevnikov, nbro_2020}.
Note that the bound of $\operatorname{tanh}$, in addition to degree normalization, essentially makes $\lim_{k \to \infty} T^{(k)}_{ij} = 0, \forall i, j$, for FAGCN (see Theorem~\ref{thm:smooth_T}).

            

\begin{table*}[t]
\begin{center}
\caption{{Node Classification Performance on Real-World Graphs}} \label{tab:performance}
    \resizebox{\textwidth}{!}{
    \setlength\tabcolsep{2.5pt}
    \renewcommand{\arraystretch}{1.1}
        \centering
        \begin{tabular}{c c c c c c c c c c c c c c}
            \toprule
            
            \vspace{0.2mm}
            \textbf{Dataset} & Chameleon & Squirrel & Actor & Texas & Cornell & Wisconsin & Computer & Photo & Wiki-CS & Pubmed & Citeseer & Cora & A.R. \\

            \midrule
            
            \textbf{Homophily}  & 0.04 & 0.03 & 0.01 & 0.00 & 0.02 & 0.05 & 0.70 & 0.77 & 0.57 & 0.66 & 0.63 & 0.77 &
            \vspace{0.2mm}
            \\
            
            \midrule
            \midrule
            \vspace{0.2mm}
            \textbf{GCN} & 67.97 $\pm$ 2.5 
                         & 53.33 $\pm$ 1.3 
                         & 30.57 $\pm$ 0.7 
                         & 65.65 $\pm$ 4.8 
                         & 58.41 $\pm$ 3.3 
                         & 62.02 $\pm$ 5.9 
                         & 81.27 $\pm$ 1.4 
                         & 90.24 $\pm$ 1.3 
                         & 79.08 $\pm$ 0.5 
                         & 79.54 $\pm$ 0.4 
                         & 72.50 $\pm$ 0.5 
                         & 83.15 $\pm$ 0.5 
                         & 9.1
                         \\
            \vspace{0.2mm}
            \textbf{APPNP}  & 53.04 $\pm$ 2.2  
                            & 40.37 $\pm$ 1.5 
                            & 35.49 $\pm$ 1.0 
                            & 79.89 $\pm$ 4.2 
                            & 80.16 $\pm$ 5.9 
                            & 84.24 $\pm$ 4.6 
                            & 81.27 $\pm$ 1.4 
                            & 91.12 $\pm$ 1.2 
                            & 79.05 $\pm$ 0.5 
                            & 79.90 $\pm$ 0.3 
                            & 73.06 $\pm$ 0.3 
                            & 83.60 $\pm$ 1.3 
                            & 7.8
                            \\

            \midrule

            \vspace{0.2mm}
            \textbf{GCN-II}  & 60.38 $\pm$ 1.9  
                                     & 48.76 $\pm$ 2.4  
                                     & 35.77 $\pm$ 1.0  
                                     & 78.59 $\pm$ 6.6  
                                     & 78.84 $\pm$ 6.6  
                                     & 83.20 $\pm$ 4.7 
                                     & 84.24 $\pm$ 1.2 
                                     & 91.81 $\pm$ 0.9  
                                     & 79.28 $\pm$ 0.6  
                                     & 80.14 $\pm$ 0.6  
                                     & {\bfseries\colorbox{green}{73.20 $\pm$ 0.8}} 
                                     & {\bfseries\colorbox{green}{85.33 $\pm$ 0.5}}
                                     & 5.5
                                     \\
            \vspace{0.2mm}
            \textbf{A-DGN}  & 69.63 $\pm$ 2.0  
                                    & 57.77 $\pm$ 1.9  
                                    & 36.41 $\pm$ 1.0  
                                    & 82.22 $\pm$ 4.8  
                                    & {\bfseries\colorbox{green}{83.14 $\pm$ 6.7}} 
                                    & {\bfseries\colorbox{green}{85.84 $\pm$ 4.0}} 
                                    & 83.70 $\pm$ 1.5  
                                    & 90.53 $\pm$ 1.3 
                                    & 79.11 $\pm$ 0.6 
                                    & 78.68 $\pm$ 0.6 
                                    & 70.16 $\pm$ 0.9 
                                    & 79.84 $\pm$ 0.9 
                                    & 6.4
                                    \\

            \midrule

            \vspace{0.2mm}
            \textbf{GAT}  & 68.01 $\pm$ 2.5 
                                  & 54.49 $\pm$ 1.7 
                                  & 30.36 $\pm$ 0.9 
                                  & 60.46 $\pm$ 6.2 
                                  & 58.22 $\pm$ 3.7 
                                  & 63.59 $\pm$ 6.1 
                                  & 84.46 $\pm$ 1.3 
                                  & 89.88 $\pm$ 1.1 
                                  & 79.44 $\pm$ 0.5 
                                  & 78.94 $\pm$ 0.4 
                                  & 71.89 $\pm$ 0.6 
                                  & 83.78 $\pm$ 0.5 
                                  & 8.5
                                  \\
            \vspace{0.2mm}
            \textbf{GATv2}  & 69.06 $\pm$ 2.2 
                                    & 57.67 $\pm$ 2.4 
                                    & 30.27 $\pm$ 0.8 
                                    & 60.32 $\pm$ 7.0 
                                    & 58.35 $\pm$ 3.8 
                                    & 61.94 $\pm$ 4.7 
                                    & 84.19 $\pm$ 1.2 
                                    & 89.87 $\pm$ 1.2 
                                    & 79.64 $\pm$ 0.5 
                                    & 79.12 $\pm$ 0.3 
                                    & 71.15 $\pm$ 1.2 
                                    & 83.88 $\pm$ 0.6 
                                    & 8.9
                                    \\
            \textbf{GATv2$^R$}  & {\bfseries\colorbox{yellow}{70.88 $\pm$ 1.9}} 
                                        & {\bfseries\colorbox{yellow}{61.23 $\pm$ 1.5}} 
                                        & 33.73 $\pm$ 0.9  
                                        & 60.68 $\pm$ 6.6  
                                        & 57.32 $\pm$ 4.5  
                                        & 60.61 $\pm$ 5.1 
                                        & 81.73 $\pm$ 2.2  
                                        & 88.71 $\pm$ 1.7 
                                        & 79.75 $\pm$ 0.6 
                                        & 78.28 $\pm$ 0.4 
                                        & 71.00 $\pm$ 0.8  
                                        & 82.42 $\pm$ 0.6 
                                        & 9.3
                                        \\
            \vspace{0.2mm}
            \textbf{GT}  & 69.34 $\pm$ 1.2 
                                 & 55.04 $\pm$ 1.9 
                                 & 36.29 $\pm$ 1.0 
                                 & {\bfseries\colorbox{yellow}{84.08 $\pm$ 5.6}} 
                                 & 80.00 $\pm$ 4.9 
                                 & {\bfseries\colorbox{yellow}{84.80 $\pm$ 4.3}}
                                 & 84.38 $\pm$ 1.3 
                                 & 91.28 $\pm$ 1.1 
                                 & {\bfseries\colorbox{yellow}{79.93 $\pm$ 0.5}} 
                                 & 79.04 $\pm$ 0.5 
                                 & 70.16 $\pm$ 0.8 
                                 & 82.09 $\pm$ 0.7 
                                 & 5.6
                                 \\
            \vspace{0.2mm}
            \textbf{FAGCN} & 60.98 $\pm$ 2.3  
                                   & 42.20 $\pm$ 1.8  
                                   & 35.67 $\pm$ 0.9 
                                   & 77.00 $\pm$ 7.7 
                                   & 78.32 $\pm$ 6.3 
                                   & 82.41 $\pm$ 3.8 
                                   & 82.79 $\pm$ 2.7  
                                   & 91.99 $\pm$ 1.0  
                                   & 79.27 $\pm$ 0.6  
                                   & 79.19 $\pm$ 0.4  
                                   & 71.55 $\pm$ 0.8  
                                   & 83.88 $\pm$ 0.5 
                                   & 7.5
                         \\
            \vspace{0.2mm}
            \textbf{DMP} & 63.79 $\pm$ 4.1 
                                 & 34.19 $\pm$ 7.6
                                 & 28.30 $\pm$ 2.7 
                                 & 66.08 $\pm$ 7.0 
                                 & 56.41 $\pm$ 5.5 
                                 & 62.73 $\pm$ 4.5 
                                 & 70.58 $\pm$ 11.3 
                                 & 82.63 $\pm$ 4.1 
                                 & 56.41 $\pm$ 7.8 
                                 & 70.07 $\pm$ 4.1 
                                 & 59.12 $\pm$ 4.4 
                                 & 75.05 $\pm$ 3.8 
                                 & 12.8
                         \\
            \midrule
            \vspace{0.2mm}
            \textbf{MixHop} & 60.30 $\pm$ 2.1 
                                    & 41.05 $\pm$ 2.0 
                                    & {\bfseries\colorbox{yellow}{36.48 $\pm$ 1.2}} 
                                    & 77.73 $\pm$ 7.3 
                                    & 75.95 $\pm$ 5.7 
                                    & 82.12 $\pm$ 4.5 
                                    & 79.52 $\pm$ 2.1 
                                    & 89.45 $\pm$ 1.5 
                                    & 78.59 $\pm$ 0.7 
                                    & 80.10 $\pm$ 0.4
                                    & 71.42 $\pm$ 0.9
                                    & 81.61 $\pm$ 0.8
                                    & 9.3
                         \\
            \vspace{0.2mm}
            \textbf{GPRGNN} & 66.92 $\pm$ 1.7
                            & 46.32 $\pm$ 1.5 
                            & 35.58 $\pm$ 0.9
                            & 81.51 $\pm$ 6.6
                            & 76.86 $\pm$ 7.1
                            & 84.06 $\pm$ 5.2
                            & 85.82 $\pm$ 0.9 
                            & {\bfseries\colorbox{yellow}{92.41 $\pm$ 0.7}}
                            & 79.67 $\pm$ 0.5 
                            & 80.28 $\pm$ 0.4 
                            & 71.59 $\pm$ 0.8
                            & 84.20 $\pm$ 0.5 
                            & \bfseries\colorbox{yellow}{5.2}
                         \\
            \vspace{0.2mm}
            \textbf{DAGNN} & 54.99 $\pm$ 2.0 
                                   & 40.03 $\pm$ 1.4 
                                   & 33.69 $\pm$ 1.0 
                                   & 61.35 $\pm$ 6.1
                                   & 63.89 $\pm$ 7.0 
                                   & 62.27 $\pm$ 4.2 
                                   & {\bfseries\colorbox{yellow}{85.83 $\pm$ 0.8}} 
                                   & 92.30 $\pm$ 0.7 
                                   & 79.31 $\pm$ 0.6 
                                   & {\bfseries\colorbox{yellow}{80.44 $\pm$ 0.5}} 
                                   & 73.16 $\pm$ 0.6  
                                   & {\bfseries\colorbox{yellow}{84.43 $\pm$ 0.5}}
                                   & 7.2
                         \\
            \midrule
            \vspace{0.2mm}
            \textbf{AERO-GNN} & {\bfseries\colorbox{green}{71.58 $\pm$ 2.4}}
                            & {\bfseries\colorbox{green}{61.76 $\pm$ 2.4}}
                            & {\bfseries\colorbox{green}{36.57 $\pm$ 1.1}}
                            & {\bfseries\colorbox{green}{84.35 $\pm$ 5.2}}
                            & {\bfseries\colorbox{yellow}{81.24 $\pm$ 6.8}}
                            & {\bfseries\colorbox{yellow}{84.80 $\pm$ 3.3}}
                            & {\bfseries\colorbox{green}{86.69 $\pm$ 1.4}}
                            & {\bfseries\colorbox{green}{92.50 $\pm$ 0.7}}
                            & {\bfseries\colorbox{green}{79.95 $\pm$ 0.5}}
                            & {\bfseries\colorbox{green}{80.59 $\pm$ 0.5}}
                            & {\bfseries\colorbox{green}{73.20 $\pm$ 0.6}}
                            & {83.90 $\pm$ 0.5} 
                            & \bfseries\colorbox{green}{1.4}
                         \\

            \bottomrule
        \end{tabular}
        }
    \begin{tablenotes}
      \item \hspace{1mm} \textbullet{ In each column, \green{$\blacksquare$} indicates ranking the first, and \yellow{$\blacksquare$} indicates ranking the second. A.R. denotes average ranking.}
    \end{tablenotes}
\end{center}
\vspace{3mm}
\end{table*}

\smallsection{Hop Attention.}
At each layer $0 \leq k \leq k_{max}$, we use $H^{(k)}$ and $\Tilde{Z}^{(k-1)}$ (for $k \geq 1$) to compute the hop attention $\Gamma^{(k)}$:
\begin{align*}
\begin{cases}
    \gamma_i^{(0)} &= (W_{hop}^{(0)})^{\top} \sigma(H_i^{(0)}) + b_{hop}^{(0)} \\
    \gamma_i^{(k)} &= (W_{hop}^{(k)})^{\top} \sigma(H_i^{(k)} || \tilde{Z}^{(k-1)}_i) + b_{hop}^{(k)}, \forall 1 \leq k \leq k_{max}
\end{cases}
\end{align*}
where $W_{hop}^{(k)}$ is a learnable weight vector and $b_{hop}$ is a learnable bias scalar.

Motivated by Theorem~\ref{thm:smooth_T},
we use node-adaptive hop attention to alleviate the problem of ``intersecting attention paths'' stated in Lemma~\ref{lem:many_long_intersections},
and we allow both positive and negative hop attention coefficients to prevent $S(T^{(k)})$ from converging to zero as $k$ goes to infinity 
(note that $\gamma_i^{(k)}$'s of the same sign within the same layer $k$ cannot change $S(T^{(k)})$).

Importantly, $b^{(k)}_{hop}$ should be initialized as 1s. This contributes significantly to model training stability by biasing the hop attention $\gamma^{(k)}_i$'s to be initialized positive.

\subsection{Theoretical Merits}
We summarize below the theoretical merits of \ours.
First, \ours provably mitigates the problems of deep graph attention stated in Section~\ref{sec:theory} (Theorems~\ref{thm:vul_to_os} and \ref{thm:smooth_T}).
\begin{theorem}\label{thm:ours_resistant}
    For \ours, $f_{att}$ is SR2OS (see Def.~\ref{def:vul_to_over_smoothing}).
\end{theorem}

\begin{theorem}\label{thm:ours_unsmooth}
    Given $G = (V, E)$ and $X$ with Assumptions~\ref{aspt:basic} and \ref{aspt:parameter_bounded} satisfied,
    for \ours, $\forall T^{(k)}$ (spec., even if $S(T^{(k)}) = 0$),
    $\exists \theta$ such that
    $S(T^{(k + 1)}) > 0$.\footnote{Recall that $\theta$ represents all the parameters in the GNN model.}
\end{theorem}

The number of parameters used by \ours is comparable to, or even smaller than, those of the edge-attention GNNs.
We ignore the parameters used in computing the first hidden features ($H^{(0)}$) and those in the output layer, since they are used in all GNN models.
Thus, we only consider the number of \textit{additional parameters}. Analysis for more models can be found in Appendix~\ref{app:complexity}.

\begin{theorem}\label{thm:number_params}
    Given the dimension $d_H$ of hidden node features and the number of layers $k_{max}$,
    for \ours and FAGCN, the number of additional parameters is $\Theta(k_{max} d_H)$;
    for GATv2, the number is $\Theta(k_{max} d_H^2)$.
\end{theorem}
\begin{proof}
    All the proofs are in Appendix~\ref{app:proofs}.
\end{proof}
    
    \section{Experiments}
    \label{sec:experiments}
    In this section, we conduct experiments to demonstrate the empirical strengths of \ours and elaborate on the theoretical analyses.

\subsection{Experimental Settings}
\label{sec:exp:settings}
\smallsection{Datasets.} 
We use 12 node classification benchmark datasets, among which 6 are homophilic and 6 are heterophilic~\cite{birds-of-a-feature, geomgcn, linkx}. In all the experiments, we use the publicly available train-validation-test splits, unless otherwise specified. We use \textit{sparse-labeled} training for homophilic graphs and \textit{dense-labeled} training for heterophilic graphs (small and large proportions of train labels, respectively; refer to Appendix~\ref{app:datasets} for details).

\smallsection{Baseline Methods.} 
The baseline methods consist of various representative attention-based GNNs,
including both \textit{edge-attention GNNs} (GAT, GATv2, GATv2$^R$, GT~\cite{gt}, FAGCN, DMP~\cite{dmp})\footnote{GATv2$^R$ is a GATv2 model with initial residual connection.}
and \textit{hop-attention GNNs} (GPRGNN, DAGNN, MixHop~\cite{mixhop}).
In addition to some \textit{simple GNNs} (GCN~\cite{gcn}, APPNP~\cite{appnp}), 
\textit{deep GNNs} without attention also serve as baselines (GCN-II~\cite{gcnii}, A-DGN~\cite{a-dgn}).
See Appendix~\ref{app:complexity} and~\ref{app:implementation-details} for their details.

\smallsection{Experiment Details.}
The Adam optimizer~\cite{adam} is used to train  the models, and the best parameters are selected based on early stopping. 
In measuring model performance (Section ~\ref{exp: model-performance}), we use 100 predetermined random seeds and report the mean $\pm$ standard deviation (SD) of classification accuracy over 100 trials. 
When analyzing attention coefficient distribution (Section ~\ref{exp: att-coef}), attention coefficients are averaged over 10 trials.

\subsection{Node-Classification Performance} \label{exp: model-performance}
We evaluate the performance of each model on 12 real-world node classification benchmarks.

\begin{figure}[t]
    \centering

    \hspace{10mm}
    \includegraphics[scale=0.35]{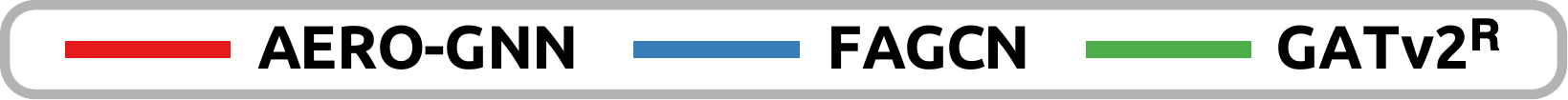}\\

    \begin{subfigure}[b]{\linewidth}
         \centering
         \includegraphics[scale=0.26]{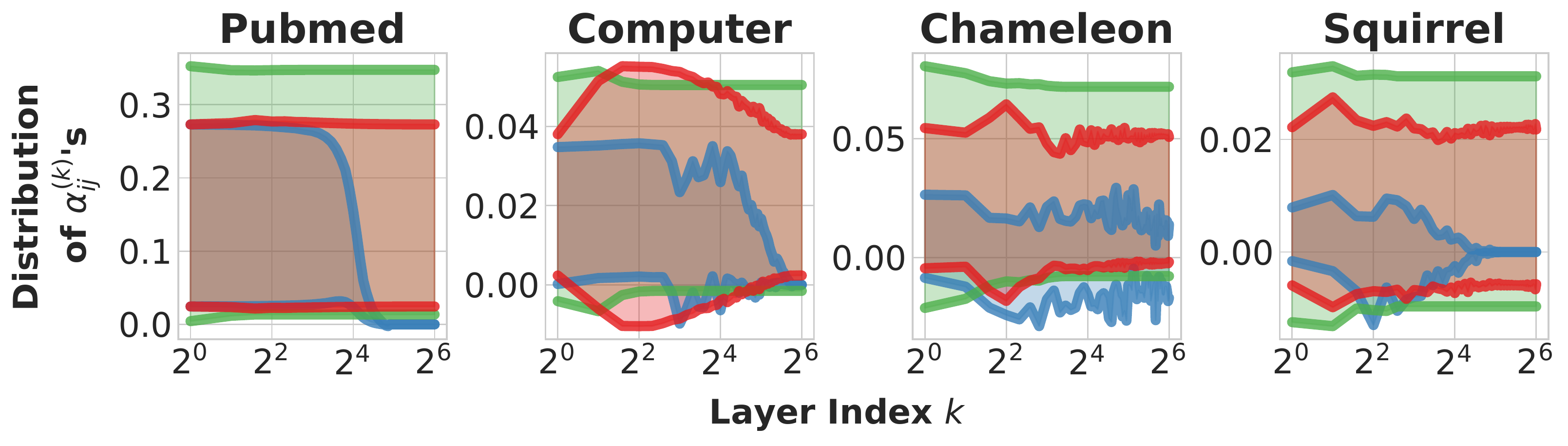}\\
         \caption{The distribution of $\alpha^{(k)}_{ij}$'s for each $k$. The shaded area represents the mean $\pm$ 1 SD.}
     \end{subfigure}
     \begin{subfigure}[b]{\linewidth}
         \centering
         \includegraphics[scale=0.26]{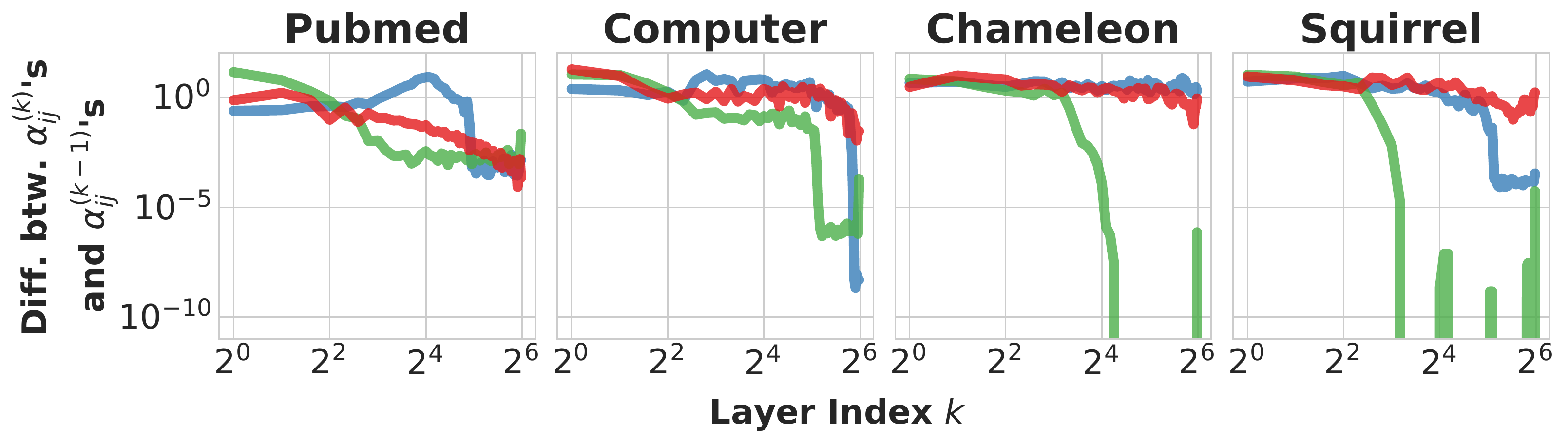}\\
         \caption{The differences between of $\alpha^{(k)}_{ij}$'s and $\alpha^{(k - 1)}_{ij}$'s for each $k$. 
         Formally, $\sqrt{\sum_{i, j} (\alpha^{(k)}_{ij} - \alpha^{(k - 1)}_{ij})^2}$ is reported for each $k$.}         
     \end{subfigure}    
    
    \caption{\label{fig:alpha}
    \bolden{Statistics of $\alpha^{(k)}_{ij}$'s for each $k$ with $k_{max}=64$.}        
    \textit{Only \ours learns edge-, hop-, and graph-adaptive edge attention over deep layers.}
    }
    \vspace{3mm} 
\end{figure}

\begin{figure}[t]
    \centering
    \hspace{5mm}
    \includegraphics[scale=0.35]{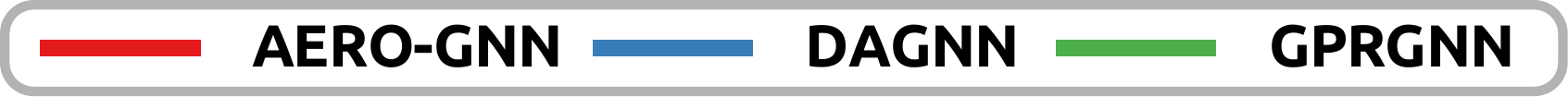}\\ 
    \begin{subfigure}[b]{\linewidth}
         \centering
         \includegraphics[scale=0.26]{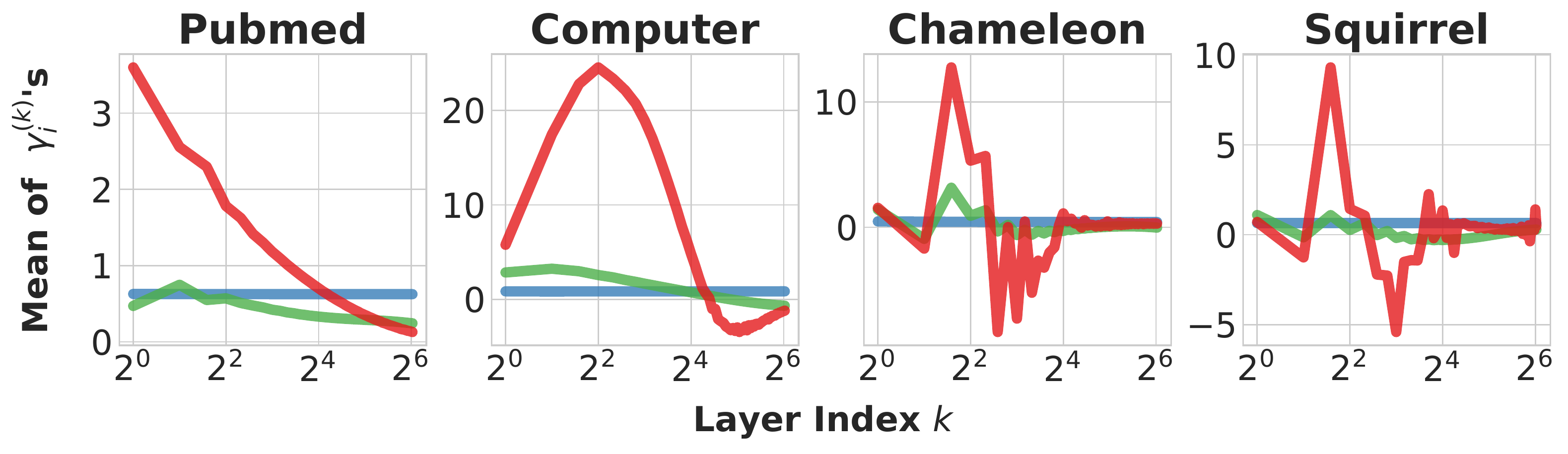}\\
         \caption{The mean value of $\gamma^{(k)}_{i}$'s for each $k$.}         
     \end{subfigure}\\
     \begin{subfigure}[b]{\linewidth}
         \centering
         \includegraphics[scale=0.26]{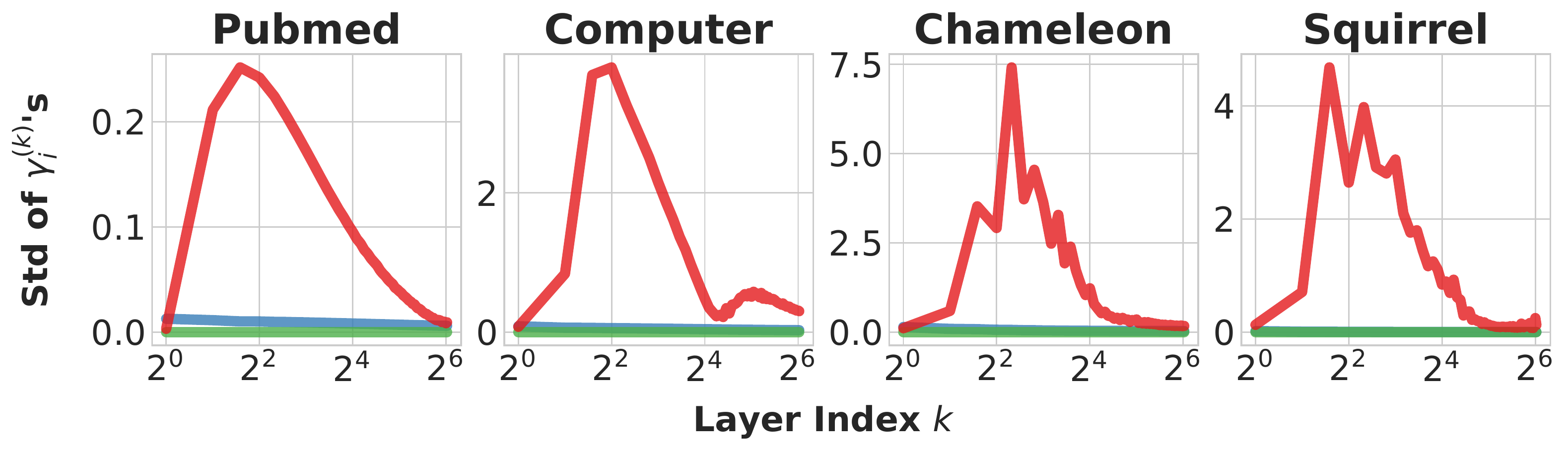}\\
         \caption{The SD of $\gamma^{(k)}_{i}$'s for each $k$.}
     \end{subfigure}    
    \caption{\label{fig:gamma}    
    \bolden{Statistics of $\gamma^{(k)}_i$'s for each $k$ with $k_{max}=64$.}    
    \textit{Only \ours learns node-, hop-, and graph-adaptive hop attention over deep layers.}
    } 
    \vspace{3mm} 
\end{figure}

\smallsection{On Heterophilic Graphs.}
As described in Section~\ref{sec:exp:settings}, dense-labeled training is used on heterophilic graphs, following prior research.
On heterophilic datasets, \ours obtains significant and consistent performance gains compared to the baseline models.
Even in relatively small graphs (e.g., \textit{texas}, \textit{cornell}, \textit{wisconsin}), where three propagation layers are enough to reach most nodes, \ours outperforms all the baseline attention-based GNNs.
In other words, even when the relative advantage of a deeper model is small, \ours still maintains competitive performance. 
Despite being specifically designed to address the heterophily problem, FAGCN, DMP, and GPRGNN are outperformed by \ours.

\smallsection{On Homophilic Graphs.}
In line with the prior research, sparse-labeled training is used for homophilic graphs.
This poses a distinct set of challenges from dense-labeled training in heterophilic graphs, especially for attention-based GNNs.
That is because, in homophilic graphs, the relative importance of each neighbor may not vary as significantly as it does in heterophilic graphs.
Additionally, with a small number of train labels, deep and complex models may prone to overfitting.
As a result, only a few models achieve strong performance in both settings (GCN-II, GT, and GPRGNN perform the best among the baselines). 
Despite such difficulties, \ours demonstrates strong performance, ranking first in 5 out of the 6 homophilic datasets.

For further evaluation, we discuss AERO-GNN's performance with multi-head attention and with datasets proposed by \citet{heterophily-eval} in Appendix~\ref{app:additional-exp}, where \ours still has the best average ranking among the competitors.

\subsection{Empirical Elaboration of the Theoretical Analysis} \label{exp: att-coef}
In this section, we empirically elaborate on the theoretical limitations of the representative attention-based GNNs (Theorems~\ref{thm:vul_to_os},~\ref{thm:smooth_T}) and show that \ours effectively mitigates the problems (Theorems~\ref{thm:ours_resistant},~\ref{thm:ours_unsmooth}).
Specifically, compared to the baselines, we show that \ours has
\begin{itemize}[leftmargin=3mm]
     \vspace{-3mm}
     \setlength\itemsep{0em}
     \item \textit{edge/node-, hop-, and graph-}adaptive attention function,
     \item \textit{less smooth} and \textit{un-smoothing} cumulative attention $T^{(k)}$, 
     \item and \textit{higher} performance at deep layers.
     \vspace{-3mm}
\end{itemize}
Here, we bring our focus back to GATv2, FAGCN, GPRGNN, DAGNN, and \ours.
\footnote{Training of vanilla GATv2 becomes unstable over deeper layers. Thus, for a fair comparison, we use GATv2$^R$ instead in the following sections.}

\smallsection{Statistics of the Attention Coefficients.}
According to Theorems~\ref{thm:vul_to_os} and \ref{thm:ours_resistant}, only \ours would have both attention functions resistant to the over-smoothed node features.
While FAGCN would only have a resistant edge-attention function, the attention coefficient distributions of the other models are expected to shrink or remain stationary with the increasing number of layers (when the over-smoothing of node features is more likely to occur). To test the hypothesis, we train 64 layers of each model and conduct post-hoc analysis of their learned attention distributions.

First, we study the edge-attention coefficients $\alpha^{(k)}_{ij}$'s.
For each $k$, Figure~\ref{fig:alpha} presents
(\textbf{a}) the distribution of $\alpha^{(k)}_{ij}$'s and
(\textbf{b}) the Frobenius norm of the difference between the attention coefficient at layer $k$ and $(k-1)$, i.e., $\norm{(\alpha^{(k)}_{ij} - \alpha^{(k-1)}_{ij})_{ij}}_F = \sqrt{\sum_{i, j} (\alpha^{(k)}_{ij} - \alpha^{(k - 1)}_{ij})^2}$.
\footnote{The layer index $k$ (the x-axis) is reversed for GATv2$^R$ and FAGCN. Refer to Appendix~\ref{app:additional-exp}.4. for rationales.}
If the distributions shrink or remain stationary over $k$ for all graphs, we have strong evidence that attention coefficients are not working properly at deep layers. 

For FAGCN, all the coefficients approach zero, and SDs of GATv2's attention coefficients remain stationary for all graphs (see Figure~\ref{fig:alpha}(a)).
Both FAGCN and GATv2's attention coefficients remain stationary at deep layers, with near 0 difference between $\alpha^{(k)}_{ij}$ and $\alpha^{(k-1)}_{ij}$ at large $k$ (see Figure~\ref{fig:alpha}(b)).

Only \ours's edge attention distributions do not shrink or remain stationary over $k$. To elaborate, \ours learns edge attention coefficients that are 
\textit{edge-adaptive} (high variances in Figure~\ref{fig:alpha}(a)), 
\textit{hop-adaptive} (high differences in Figure~\ref{fig:alpha}(b)),
and \textit{graph-adaptive} (diverse patterns for different graphs in Figure~\ref{fig:alpha}(a)).
The results for all 12 datasets are in Appendix~\ref{app:additional-exp}.

We now investigate hop-attention coefficients $\gamma^{(k)}_i$'s.
Figure~\ref{fig:gamma}(a) shows the mean value
and (b) the SD of $\gamma_i^{(k)}$'s, for each layer $k$.
Each figure respectively suggests how hop-adaptive and node-adaptive $\gamma_i^{(k)}$'s are. 

Again, as expected, hop attention distributions of DAGNN remain stationary over deep layers, and hence, they are less adaptive to node, hop, or graph 
(see stationary values in Figure~\ref{fig:gamma}(a) and the very small SD values in Figure~\ref{fig:gamma}(b)).
Since GPRGNN's hop attention $\gamma_i^{(k)}$'s are free parameters, it learns hop-adaptive and graph-adaptive $\gamma_i^{(k)}$'s regardless of over-smoothing. 
However, they are not node-adaptive.

In stark contrast, hop attention coefficients of \ours are 
\textit{node-adaptive} (high SD in Figure~\ref{fig:gamma}(b)), 
\textit{hop-adaptive} (mean value changes over different layers in Figure~\ref{fig:gamma}(a)),
and \textit{graph-adaptive} (diverse patterns for different graphs in Figure~\ref{fig:gamma}(a) and (b)). 
The results for all 12 datasets are in Appendix~\ref{app:additional-exp}.

Through this series of empirical analyses, we present strong evidence that attention functions of the representative attention-based GNNs, except for those of \ours, are vulnerable to node feature over-smoothing and fail to remain expressive over deep layers.

\begin{figure}[t]
    \centering    
    \includegraphics[scale=0.35]{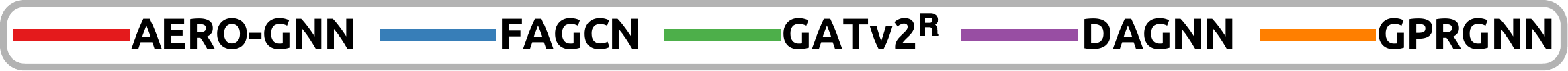}\\
    \begin{subfigure}[b]{\linewidth}
         \centering
         \includegraphics[scale=0.26]{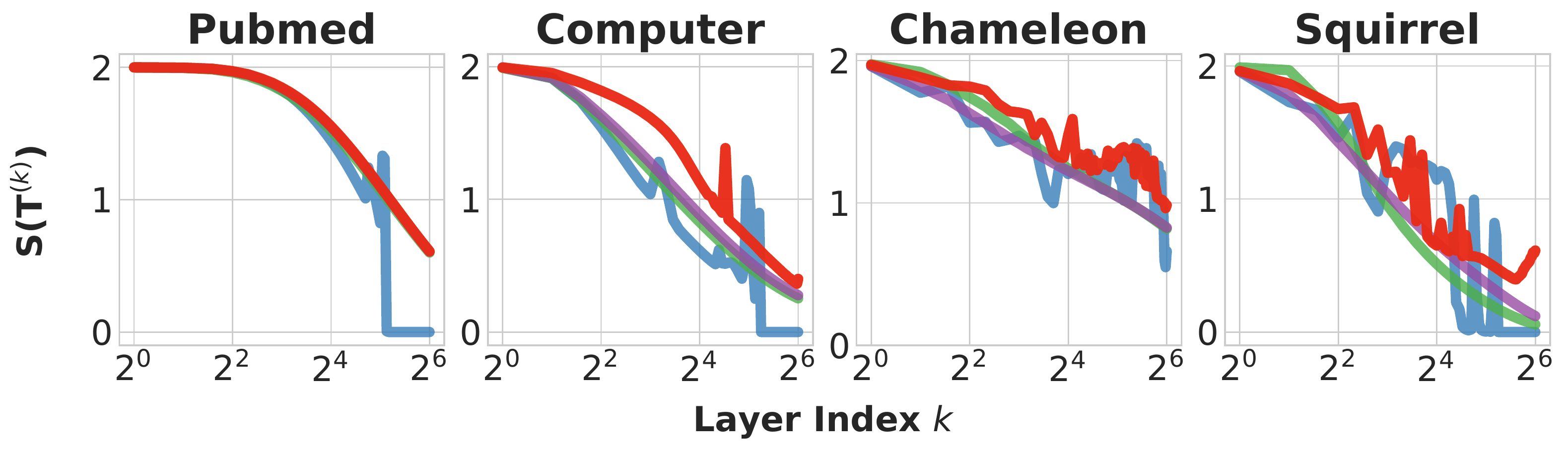}\\
         \caption{The smoothness score $S(T^{(k)})$ of the cumulative attention matrix $T^{(k)}$ for each $k$.}         
     \end{subfigure}
    \begin{subfigure}[b]{\linewidth}
         \centering
         \includegraphics[scale=0.26]{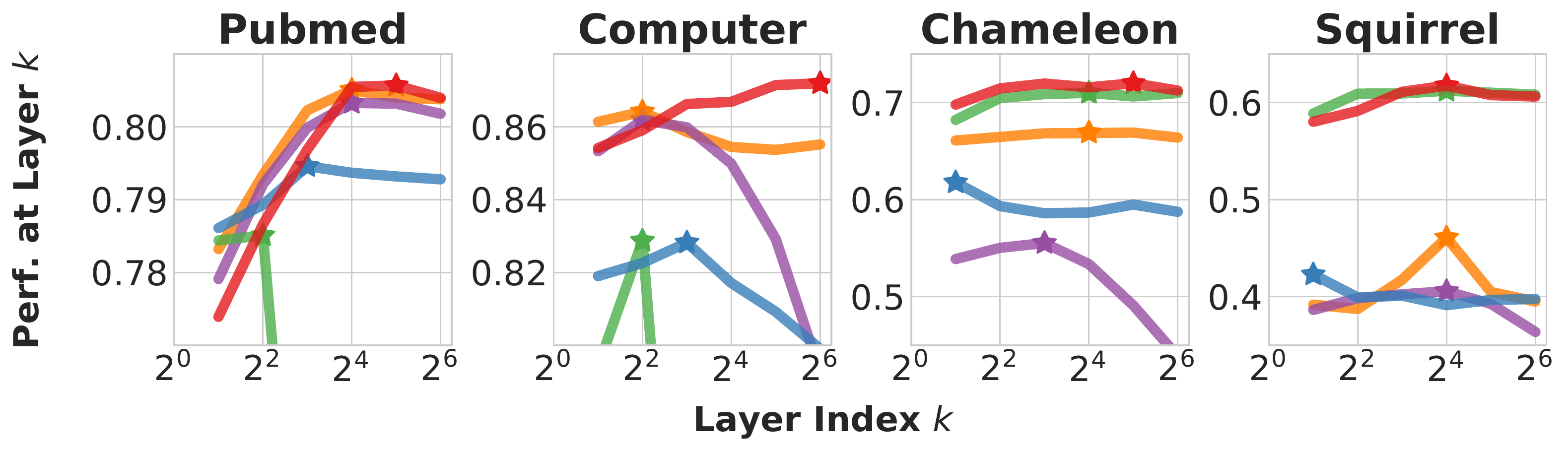}\\
         \caption{The model performance at each layer $k \in \set{2,4,8,16,32,64}$. A star ($\filledstar$) indicates best performance of each model.}
     \end{subfigure}     
    \caption{\label{fig:st_perf}
    \bolden{The smoothness of the cumulative attention matrix $T^{(k)}$ and the model performance.}
    \textit{\ours has (\textbf{a}) less smoothed $T^{(k)}$ over deep layers and
    (\textbf{b}) higher best performance achieved at a deeper layer.}
    } 
\end{figure}

\smallsection{Cumulative Attention and Model Performance.}
According to Theorems~\ref{thm:smooth_T} and \ref{thm:ours_unsmooth}, only \ours can avoid completely-smoothed cumulative attention (i.e. $S(T^{(k)})=0$) over deep layers, highlighting its capacity to learn meaningful attention at any model depth. Here, we empirically elaborate on the theoretical analysis.

Figure ~\ref{fig:st_perf}(a) shows the smoothness scores of cumulative attention matrices $S(T^{(k)})$.
\footnote{Recall that a lower $S(T^{(k)})$ indicates that the cumulative attention $T^{(k)}$ is more smoothed} 
\ours generally has less smooth $T^{(k)}$ over deep layers.
Notably, there often occurs un-smoothing of $T^{(k)}$, such that $S(T^{(k)}) > S(T^{(k-1)})$, only for \ours and FAGCN.
This phenomenon is attributable to the use of negative attention in both models.
However, $S(T^{(k)})$ of FAGCN quickly converges to 0. These findings resonate with Theorem ~\ref{thm:ours_unsmooth}, showing that \ours's attention function does remain expressive at deep layers. The results for all 12 datasets are in Appendix~\ref{app:additional-exp}.
\footnote{If $S(T^{(k)}) \leq S(T^{(k-1)})$ always holds, then $S(T^{(k)})$ is convergent. See Corollary~\ref{cor:decreasing_then_conv} in Appendix~\ref{app:proofs}.} 
\begin{table}[t]
\begin{center}
\caption{{Best Performing Layers Used in Table~\ref{tab:performance}}} \label{tab:best-layer}
    \resizebox{\linewidth}{!}{
    \setlength\tabcolsep{3pt}
    \renewcommand{\arraystretch}{1.0}
        \centering
        \begin{tabular}{c c c c c c c c c c c c c}
            \toprule
            
            \textbf{Dataset} & CM 
                            & SQ 
                            & AT
                            & TX
                            & CN 
                            & WC
                            & CP 
                            & PT
                            & WK 
                            & PM 
                            & CS 
                            & CR \\
            
            \midrule
            \midrule
            
            \textbf{GCN-II}  & 8
                                     & 4
                                     & 4
                                     & 4
                                     & 4
                                     & 4
                                     & 4
                                     & {\textbf{32}}
                                     & 4
                                     & 16
                                     & {\textbf{32}}
                                     & {\textbf{64}}
                                     \\
                         
            \textbf{A-DGN}  & 16
                                    & {\textbf{32}}
                                    & {\textbf{32}}
                                    & 2
                                    & {\textbf{32}}
                                    & 2
                                    & 8
                                    & 8
                                    & 16
                                    & 8
                                    & 16
                                    & 4
                                    \\
            \midrule
            \midrule
            \textbf{GATv2$^R$}  & 16
                                        & 16
                                        & 16
                                        & \textbf{8}
                                        & 2
                                        & 2
                                        & 4
                                        & 4
                                        & 2
                                        & 4
                                        & 2
                                        & 2
                                        \\

            \textbf{FAGCN} & 2
                                   & 2
                                   & 6
                                   & 7
                                   & 2
                                   & 7
                                   & 7
                                   & 8
                                   & 3
                                   & 8
                                   & 4
                                   & 4
                         \\
            \textbf{GPRGNN} & 16
                                   & 16
                                   & 4
                                   & 4
                                   & 4
                                   & \textbf{8}
                                   & 4
                                   & 4
                                   & {\textbf{32}}
                                   & 16
                                   & 8
                                   & 8
                         \\

            \textbf{DAGNN} & 5
                                   & 20
                                   & 5
                                   & 5
                                   & 5
                                   & 5
                                   & 5
                                   & 10
                                   & 5
                                   & 20
                                   & 10
                                   & 10
                         \\
                         
            \midrule
            \textbf{AERO-GNN} & {\textbf{32}}
                            & 16
                            & 4
                            & {\textbf{8}}
                            & 4
                            & 4
                            & {\textbf{32}}
                            & {\textbf{32}}
                            & 16
                            & {\textbf{32}}
                            & {\textbf{32}}
                            & 32
                         \\

            \bottomrule
        \end{tabular}
        }
\end{center}
\vspace{3mm}
\end{table}

Figure ~\ref{fig:st_perf}(b) illustrates the model performance at layer $k \in \set{2,4,8,16,32,64}$. \ours generally achieves better performance over deeper layers. Meanwhile, the performance of the representative attention-based GNNs often drops, even significantly, at deep layers. Table~\ref{tab:best-layer} further shows that \ours generally achieves its best performances at deeper layers than the attention-based GNNs.

In this section, we empirically evaluated the adaptiveness of attention coefficients, smoothness of $T^{(k)}$, and performance with depth for each model.
This set of empirical observations, in concert with the theoretical findings, indicate that \ours indeed learns the most expressive deep graph attention.

    \section{Discussion}
    \label{sec:discussion}
    Graph attention has had a significant impact on the research and applications of GNNs. 
Many attention-based GNNs have been designed, investigating how to better infer relations between node pairs. 
Some models introduced more features or loss terms~\cite{supergat, cgat, cat}, and some others relaxed their attentions to include negative values~\cite{fagcn, gprgnn, dmp}. 

On the other hand, making GNNs deeper to enhance their expressivity has been an important setback to GNN research. 
A number of problems have been proposed to limit its expressiveness to increase over depth. 
Not to mention over-smoothing, but also over-squashing and over-correlation have been pointed out as possible problems in building deep GNNs. 
As such, techniques that modulate propagation or aggregation functions~\cite{grand, a-dgn, deepergcn, sheaf}, residual connection functions~\cite{can_gcn_go_cnn, gcnii, 1000layers}, hidden features~\cite{groupnorm, pairnorm, contrast-norm}, or graph topologies~\cite{dropedge, oversmoothing_natural, shadow-gnn, curvature, sheaf} have been applied to build deeper GNNs.

In this work, we bridge the two research directions, addressing two underexplored questions:
(\textbf{a}) what are the unique challenges in deep graph attention, and 
(\textbf{b}) how can we design provably more expressive deep graph attention? 
We argue that the representative attention-based GNNs suffer from the proposed set of problems, possibly on top of the general problems in deep GNNs. 
Thus, we design \ours to theoretically and empirically mitigate the problems. 

Under a larger context, these findings extend prior literature on limitations to \textit{deep attention in general}.
Specifically, similar problems of deep attention smoothing have been reported for transformers~\cite{transformer, attention-is-not-all-you-need}, in both natural language processing~\cite{bert-over-smoothing} and computer vision~\cite{vit-smoothing, deepvit} domains. 
We demonstrate that attention-based GNNs share related, yet distinct, problems and propose a novel solution.
Hence, we expect this work will inspire future research on deep attention and graph learning in various directions. 

The generalizability of the present work is limited in that linear propagation is assumed to define $T^{(k)}$ of GATs. 
Also, we deliberately suppress non-linearity in the \textit{propagation layers} of AERO-GNN to focus on the ability to learn dynamic receptive fields, expressed in $T^{(k)}$. 
Still, it can be a promising future work to add non-linear propagation to \ours to address other challenges and applications to more complex tasks.

    \section*{Acknowledgements}
    This work was supported by Institute of Information \& Communications Technology Planning \& Evaluation (IITP) grant funded by the Korea government (MSIT) (No. 2022-0-00871, Development of AI Autonomy and Knowledge Enhancement for AI Agent Collaboration) (No. 2019-0-00075, Artificial Intelligence Graduate School Program (KAIST)).
    
    \normalem
    \bibliography{ref.bib}
    \bibliographystyle{icml2023}

    \clearpage    
    \appendix
    \appendix

\section{Proofs and Additional Theoretical Results}\label{app:proofs}
In this section, we provide proofs of the theoretical claims in the main text and some additional theoretical results.

\subsection{Regarding Problem 1}

\begin{proof}[Proof of Theorem~\ref{thm:vul_to_os}]
    We analyze $\tilde{\calA}^{(k)}$'s and $\Gamma^{(k)}$'s for each considered GNN model.
    
    \smallsection{GATv2.} For GATv2, for all $i, j, k$,
    \[
    \tilde{\alpha}^{(k)}_{ij} = \operatorname{exp}((W^{(k)}_{edge})^{\top} \sigma(H^{(k-1)}_i || H^{(k-1)}_j)),
    \]
    and $\gamma^{(k)}_i \equiv 1$.
    It is easy to see that $\tilde{\alpha}^{(k)}_{ij}$ is totally determined by $H^{(k)}_i$ and $H^{(k)}_j$,
    and $\gamma^{(k+1)}_{i}$ is constant, for all $i, j$.
    Thus, if the conditions in Definition~\ref{def:vul_to_over_smoothing} hold, then for any fixed $\theta$,
    $\tilde{\alpha}^{(k)}_{ij} = \tilde{\alpha}^{(k)}_{i'j'}$ and $\gamma^{(k+1)}_{i} = \gamma^{(k+1)}_{i'}$
    Hence, for GATv2, $f_{att}$ is V2OS.
    
    \smallsection{GAT Variants.} Several variants of GAT are worth mentioning.
    CATs~\cite{cat} use structural features to compute the attention coefficients at each layer. 
    This enables CATs to partially overcome the limitation of the original GAT, and
    it is easy to see that $f_{att}$ is WR2OS but not SR2OS (since CATs only essentially use identical hop attention).
    SuperGAT~\cite{supergat}, another GAT variant, further uses self-supervision for its attention function.
    However, the attention function of SuperGAT is still completely dependent on the node features in the current layer, and thus 
    for SuperGAT, $f_{att}$ is V2OS as GATv2.
    
    \smallsection{FAGCN.} For FAGCN, for all $i, j, k$,
    \[
    \Tilde{\alpha}^{(k)}_{ij} = \operatorname{tanh}((W^{(k)}_{edge})^{\top}(Z^{(k-1)}_i || Z^{(k-1)}_j)),
    \]
    and
    $\gamma^{(k)}_i \equiv \gamma_0$, for some constant $\gamma_0$.
    If the conditions in Definition~\ref{def:vul_to_over_smoothing} hold, then for any fixed $\theta$,
    Since $X_i \neq X_j, \forall i, j$,
    there exist $W^{(1)}$ (and $W^{(2)}$) such that
    $H^{(0)}_{i}$, $H^{(0)}_{j}$, $H^{(0)}_{i'}$, $H^{(0)}_{j'}$ are all distinct.
    $H^{(0)}$ is used to compute $Z^{(k)}$ for each $k$,    
    and $Z^{(k)}$ is used to compute $\calA^{(k + 1)}$ for each $k$.
    Thus, even if the conditions in Definition~\ref{def:vul_to_over_smoothing} hold,
    there still exists $\theta$ such that
    $\tilde{\alpha}^{(k)}_{ij} \neq \tilde{\alpha}^{(k)}_{i'j'}$, 
    but
    $\gamma^{(k+1)}_{i} = \gamma^{(k+1)}_{i'}$.
    Hence, for FAGCN, $f_{att}$ is WR2OS.
    
    \smallsection{GPRGNN.} For GPRGNN, for all $i, j, k$,
    $\tilde{\alpha}^{(k)}_{ij} \equiv 1$, and
    $\gamma^{(k)}_i \equiv \gamma_0$, for some learnable scalar $\gamma_0$.
    Hence, it is easy to see that for GPRGNN,
    $f_{att}$ is V2OS.
    
    \smallsection{DAGNN.} For DAGNN, for all $i, j, k$,
    $\tilde{\alpha}^{(k)}_{ij} \equiv 1$, and
    \[
    \gamma^{(k)}_i = \operatorname{sigmoid}((W^{(k)}_{hop})^{\top} H^{(k)}_i).
    \]
    With similar reasoning for GATv2,
    it is easy to see that for DAGNN,
    $f_{att}$ is also V2OS.   
\end{proof}

\begin{proof}[Proof of Theorem~\ref{thm:ours_resistant}] We now prove how \ours mitigates problem 1.

    \smallsection{\ours.} For \ours, for all $i, j$ and $k>0$,
    \[
    \tilde{\alpha}^{(k)}_{ij} = \operatorname{softplus}((W^{(k)}_{edge})^{\top} \sigma(\tilde{Z}^{(k-1)}_i || \tilde{Z}^{(k-1)}_j)),
    \]
    and
    \[
    \gamma_i^{(k)} = (W_{hop}^{(k)})^{\top} \sigma(H_i^{(k)} || \tilde{Z}^{(k-1)}_i) + b_{hop}^{(k)}.
    \] 
    Since $X_i \neq X_j, \forall i, j$,
    there exist $W^{(1)}$ (and $W^{(2)}$ when $W^{(2)}$ is also used) such that
    $H^{(0)}_{i}$, $H^{(0)}_{j}$, $H^{(0)}_{i'}$, $H^{(0)}_{j'}$ are all distinct.
    $H^{(0)}_i$ is used to compute $\gamma^{(0)}_i$ for each $i$.
    $\Gamma^{(0)}$ is used to compute $Z^{(k)}$ for each $k$,
    and $Z^{(k)}$ is used to compute $\Gamma^{(k + 1)}$ for each $k$.
    Thus, even if the conditions in Definition~\ref{def:vul_to_over_smoothing} hold,
    there still exists $\theta$ such that
    $\tilde{\alpha}^{(k)}_{ij} \neq \tilde{\alpha}^{(k)}_{i'j'}$ and
    $\gamma^{(k+1)}_{i} \neq \gamma^{(k+1)}_{i'}$.
\end{proof}

\subsection{Regarding Problem 2}

\begin{lemma}\label{lem:sm_bounded}
    $S(T)$ is bounded, $\forall T \in \bbR^{n \times n}$.
\end{lemma}
\begin{proof}
    A lower bound $0$ is straightforward.
    Regarding an upper bound, for any $T$,
    \begin{align*}
    S(T) &= (\sum_{i, j \in \binom{[n]}{2}} \norm{\frac{T_i}{\norm{T_i}}_1 - \frac{T_j}{\norm{T_j}_1}}_1) / \binom{n}{2} \\
    &\leq (\sum_{i, j \in \binom{[n]}{2}} \norm{\frac{T_i}{\norm{T_i}}_1}_1 + \norm{\frac{T_j}{\norm{T_j}}_1}_1) / \binom{n}{2}\\
    &\leq 2
    \end{align*}
\end{proof}

\begin{corollary}\label{cor:decreasing_then_conv}
    For an infinite series $\set{T_k}_{k = 0}^{\infty}$,
    if $S(T_{k + 1}) < S(T_k), \forall k$,
    then $S(T_k)$ is convergent, i.e.,
    $\lim_{k \to \infty} S(T_k)$ exists and is finite.
\end{corollary}
\begin{proof}
    It is straightforward by the monotone convergence theorem.
\end{proof}

\begin{lemma}\label{lem:zero_sm_iff_rank1}
    For a matrix $T$ with at least one non-zero entry, $S(T) = 0$ if and only if $\operatorname{rank}(T) \leq 1$.
\end{lemma}
\begin{proof}
    It is easy to see that
    $T_i / \norm{T_i}_1$ are all identical for all $i$, 
    if $\operatorname{rank}(T) \leq 1$;
    and $\exists i \neq j, T_i / \norm{T_i}_1 \neq T_j / \norm{T_j}_1$, 
    if $\operatorname{rank}(T) \geq 2$.
\end{proof}
\begin{remark}
    Lemmas~\ref{lem:sm_bounded} and \ref{lem:zero_sm_iff_rank1} also hold when we use other norms in $S(\cdot)$.
\end{remark}

\begin{proof}[Proof of Theorem~\ref{thm:smooth_T}]
    We analyze $S(T^{(k)})$ as $k \to \infty$ for each model.
    
    \smallsection{GATv2.} For GATv2, for each $k$, $\Gamma^{(k)} = I$ and thus $T^{(k)} = \prod_{\ell= 1}^{k} \calA^{(\ell)}$, where
    $\alpha^{(\ell)}_{ij} = \frac{\tilde{a}^{(\ell)}_{ij}}{\sum_{j' \in N(i)} \tilde{a}^{(\ell)}_{ij'}} \in (0, 1), \forall i, j$
    ({see Table~\ref{tab:attention_functions}}).
    Specifically, $\sum_{j} \alpha^{(\ell)}_{ij} = 1, \forall i, \ell$, 
    i.e., $\calA^{(\ell)}$ is row-wise stochastic, for each $\ell$.
    Since the product of two row-wise stochastic matrices is again still row-wise stochastic, 
    $T^{(\ell)}$ is also row-wise stochastic, for each $\ell$.    
    Fix any $k$, we have
    $T^{(k + 1)} = T^{(k)} \calA^{(k + 1)}$,
    and
    \begin{align*}
        S(T^{(k + 1)}) &= \sum_{i, j} \norm{\frac{T^{(k + 1)}_i}{\norm{T^{(k + 1)}_i}_1} - \frac{T^{(k + 1)}_j}{\norm{T^{(k + 1)}_j}_1} }_1 \\
        &= \sum_{i, j} \norm{T^{(k + 1)}_i - T^{(k + 1)}_j}_1 \\        
        &= \sum_{i, j} \sum_x \abs{T^{(k + 1)}_{ix} - T^{(k + 1)}_{jx}} \\
        &= \sum_{i, j} \sum_x \abs{\sum_{y \in N(x)} T^{(k)}_{iy} \calA^{(k+1)}_{yx} - \sum_{y \in N(x)} T^{(k)}_{jy} \calA^{(k+1)}_{yx}} \\
        &= \sum_{i, j} \sum_x \abs{\sum_{y \in N(x)} (T^{(k)}_{iy} -T^{(k)}_{jy}) \calA^{(k+1)}_{yx}} \\        
        &\leq \sum_{i, j} \sum_x \sum_{y \in N(x)} \abs{T^{(k)}_{iy} -T^{(k)}_{jy}} \abs{\calA^{(k+1)}_{yx}} \\                
        &= \sum_{i, j} \sum_x \sum_{y \in N(x)} \abs{T^{(k)}_{iy} -T^{(k)}_{jy}} \calA^{(k+1)}_{yx} \\ 
        &= \sum_{i, j} \sum_y \abs{T^{(k)}_{iy} -T^{(k)}_{jy}} \sum_{x \in N(y)} \calA^{(k+1)}_{yx} \\ 
        &= \sum_{i, j} \sum_y \abs{T^{(k)}_{iy} -T^{(k)}_{jy}} = S(T^{(k)}),
    \end{align*}
    where the inequality holds as equality if and only if
    $T^{(k)}_{i} = T^{(k)}_{j}, \forall i, j$, which means
    $S(T^{(k)}) = 0$ and thus $S(T^{(k')})$ remains $0$ for all $k' \geq k$.
    \footnote{Here, we ignore $(/ \binom{n}{2})$ in computing $S(T^{(k)})$ for simplicity.}
    Otherwise, the inequality holds as a strict inequality, i.e., $S(T^{(k + 1)}) < S(T^{(k)}), \forall k$, which, by Corollary~\ref{cor:decreasing_then_conv}, implies that $S(T^{(k)})$ converges to a constant as $k$ goes to infinity.    
    In conclusion, either $S(T^{(k)}), \forall k \geq k'$ remains zero after some $k'$, or it strictly decreases and converges and as $k$ goes to infinity.
    
    Moreover, by Theorem~$3$ in~\citet{chen2016convergence}, and Assumption~\ref{aspt:basic} (specifically, $G$ is connected and non-bipartite),
    if there exists $c'' > 0$ s.t. for all $k$, each nonzero entry of $\calA^{(k)}$ is at least $c''$,
    then $T^{(k)} = \prod_{\ell= 1}^{k} \calA^{(\ell)}$ converges to a matrix with equal rows, which implies that $S(T^{(k)})$ converges to $0$.
    Indeed,
    by Assumption~\ref{aspt:parameter_bounded},
    for all $i, j, k$,
    $\tilde{\alpha}^{(k)}_{ij} = \exp((W^{(k)}_{edge})^\top \sigma(H^{(k - 1)}_i || H^{(k - 1)}_j))$, where
    $(W^{(k)}_{edge})^\top \sigma(H^{(k - 1)}_i || H^{(k - 1)}_j) \in [-nc^2_{param}, nc^2_{param}]$, and thus    
    each nonzero $\alpha^{(k)}_{ij}$ satisfies that
    \[
    \alpha^{(k)}_{ij} \geq \frac{\exp(-nc^2_{param})}{\exp(-nc^2_{param}) + (n-1)\exp(nc^2_{param})}.
    \]    
    We complete the proof for GATv2 by letting 
    \[
    c'' = \frac{\exp(-nc^2_{param})}{\exp(-nc^2_{param}) + (n-1)\exp(nc^2_{param})}.
    \]
    \smallsection{GAT Variants.} The proof for GATv2 is mainly based on (\textbf{a}) trivial hop attention function and (\textbf{b}) the row-wise stochasticity introduced by the $\operatorname{softmax-}$based normalization. Trivial hop attention and $\operatorname{softmax}$ normalization scheme are also used in other variants, including SuperGAT~\cite{supergat} and CATs~\cite{cat}. Therefore, for SuperGAT and CATs, the conclusion is the same as for GATv2, i.e., $\lim_{k \to \infty} S(T^{(k)}) = 0$.

    \smallsection{FAGCN.} For FAGCN, we have $\forall i, j, k,
    \alpha^{(k)}_{ij} = \tilde{\alpha}^{(k)}_{ij} / \sqrt{d_i d_j}$, where
    $\tilde{\alpha}^{(k)}_{ij} \in (-1, 1)$ due to the usage of $\operatorname{tanh}(\cdot)$ ({see Table~\ref{tab:attention_functions}}),
    and $\Gamma^{(k)}$ ($\gamma_i^{k}$) is a constant, and we let $\Gamma$ ($\gamma$) denote it.  
    Let $\norm{\cdot}_2$ and $\norm{\cdot}_F$ denote the spectral norm and the Frobenius norm, respectively.
    For any fixed $k_0$, $\norm{Z^{(k_0)}}_2 \leq \norm{Z^{(k_0)}}_F \leq n\sqrt{c_{param}}$,
    and thus
    \begin{align*}
        \Tilde{\alpha}^{(k_0+1)}_{ij} &\leq \operatorname{tanh}((W^{(k_0 + 1)}_{edge})^{\top} (Z^{(k_0)}_i || Z^{(k_0)}_j) \\
        &\leq \operatorname{tanh} (2\sqrt{n} c_{param} \norm{Z^{(k_0)}}_2) \\
        &\leq \operatorname{tanh} (2\sqrt{n} c_{param} nc_{param}) \\
        &= \operatorname{tanh} (2n^{1.5} c_{param}^2).
    \end{align*}
    Therefore, the spectral norm of $\calA^{(k_0)}$ is
    smaller than that of $c_{sn}\Tilde{A}$ (see the proof for GPRGNN and DAGNN), where $c_{sn} = \operatorname{tanh} (2n^{1.5} c_{param}^2)$.
    Specifically,
    \[
    \norm{\calA^{(k)}}_2 < \norm{c_{sn}\Tilde{A}}_2 \leq c_{sn} \norm{\Tilde{A}}_2 \leq c_{sn},
    \]
    which, by the submultiplicity of the spectral norm, gives $\norm{\prod_{\ell=1}^k \calA^{(\ell)}}_2 \leq {c_{sn}^k}$,
    and thus
    $\lim_{k \to \infty} \norm{\prod_{\ell=1}^k \calA^{(\ell)}}_2 = 0$.
    Since $\Gamma^{(k)}$ is bounded,
    $\lim_{k \to \infty} \norm{T^{(k)}}_2 = 0$, by which we conclude that
    $\lim_{k \to \infty} T^{(k)}_{ij} = 0, \forall i, j$, completing the proof of FAGCN.
    
    \smallsection{GPRGNN and DAGNN.} For GPRGNN and DAGNN, $\calA^{(k)} = \Tilde{A}, \forall k$, where 
    $\Tilde{A} = (D + I)^{-1/2} (A + I) (D + I)^{-1/2}$,
    and thus 
    \begin{align*}
        T^{(k)} &= \Gamma^{(k)} \prod_{\ell= k}^{1} \calA \\
        &= \Gamma^{(k)} \prod_{\ell= k}^{1} \Tilde{A} \\
        &= \Gamma^{(k)} \Tilde{A}^k \\
        &= \Gamma^{(k)} (D+I)^{-1/2} ((A+I) (D+I)^{-1})^{-k + 1} (A+I) (D+I)^{-1/2}.
    \end{align*}
    Since $G$ is connected and non-bipartite,
    by the Ergodic theorem on Markov chains,
    $(D + I)^{-1/2} ((A + I) (D + I)^{-1})^{-k + 1}$ converges to a rank-one matrix (let $P_T$ denote this rank-one matrix) as $k \to \infty$.
    Furthermore, since $(A + I) (D + I)^{-1/2}$ is bounded for any given $G$, formally, $\forall \epsilon < 0, \exists K > 0, k \geq K \Rightarrow \norm{\Tilde{A}^k - P_T (A + I) (D + I)^{-1/2}} \leq \epsilon$,
    where $\operatorname{rank}(P_T (A + I) (D + I)^{-1/2}) \leq \operatorname{rank}(P_T) = 1$.
    Since $S$ is continuous, and $S(P_T (A + I) (D + I)^{-1/2}) = 0$ (see Lemma~\ref{lem:zero_sm_iff_rank1}), $S(\Tilde{A}^k)$ converges to $0$ as $k \to \infty$.
    Since $\Gamma^{(k)}$ is diagonal and the signs of all entries are identical for both DAGNN and GPRGNN, $S(\Gamma^{(k)} \Tilde{A}^k) = S(\Tilde{A}^k)$, completing the proof for GPRGNN and DAGNN.

\end{proof}

\begin{proof}[Proof of Theorem~\ref{thm:ours_unsmooth}]
    We now prove how \ours mitigates problem 2.
    
    \smallsection{\ours.} For \ours, fix any $k$, consider first
    $S(\Bar{T}^{(k)})$, where $\Bar{T}^{(k)} = \prod_{\ell= 1}^{k} \calA^{(\ell)}$.
    If $S(\Bar{T}^{(k)}) > 0$, then
    {simply letting $W^{(k)}_{hop}$ be a zero vector and 
    letting $b^{(k)}_{hop} = 1$ suffices},
    since in that case $\Gamma^{(k)} = I$, and
    $S(T^{(k)}) = S(\prod_{\ell= 1}^{k} \calA^{(\ell)}) > 0$.
    Otherwise, if $S(\prod_{\ell= 1}^{k} \calA^{(\ell)}) = 0$,
    i.e.,
    $T_i / \norm{T_i}_1 = T_j / \norm{T_j}_1, \forall i, j$,
    by the formulae of $\Gamma$,
    we claim that $\exists \theta$ s.t. $\gamma^{(k)}_i \gamma^{(k)}_{i'} < 0$,
    for at least one pair $i, j$.
    This is true since as shown in the proof of Theorem~\ref{thm:ours_unsmooth}, we can find a pair $i, j$ such that $\sigma(H_i^{(k)} || \tilde{Z}^{(k-1)}_i) \neq \sigma(H_j^{(k)} || \tilde{Z}^{(k-1)}_j)$, and thus we can find $W^{(k)}_{hop}$ and $b^{(k)}_{hop}$ such that $\gamma^{(k)}_i \gamma^{(k)}_{i'} < 0$.
    Specifically, WLOG, suppose that the $t$-th entry of $\sigma(H_i^{(k)} || \tilde{Z}^{(k-1)}_i)$ and $\sigma(H_j^{(k)} || \tilde{Z}^{(k-1)}_j)$ are different, then we can let $W^{(k)}_{hop}$ be the unit vector with $(W^{(k)}_{hop})_t = 1$ and all the other entries being zero, and let $b^{(k)}_{hop} = -((\sigma(H_i^{(k)} || \tilde{Z}^{(k-1)}_i))_t + (\sigma(H_j^{(k)} || \tilde{Z}^{(k-1)}_j))_t) / 2$.    
    Thus, 
    \begin{align*}
        S(T^{(k)}) \geq \norm{ \frac{\gamma^{(k)}_i T_i}{\norm{\gamma^{(k)}_i T_i}_1} - \frac{\gamma^{(k)}_{i'} T_j}{\norm{\gamma^{(k)}_{i'} T_j}_1} } = 2\norm{ \frac{T_i}{\norm{T_i}_1}} > 0,
    \end{align*}
    completing the proof.\footnote{Moreover, since we use edge attention (compared to GPRGNN and DAGNN),
    and we have control over the magnitude of $\Gamma^{(k)}$ (compared to all the others), we can make sure that $T^{(k)}$ does not converge to an all-zero matrix.}
\end{proof}

\begin{proof}[Proof of Lemma~\ref{lem:many_long_intersections}]
    Since $G$ is non-bipartite, it contains a cycle of odd length.
    WLOG, suppose such a cycle $\mathbf{c}$ includes a node $y \in V$ (possibly, $y \in \set{i, j, x}$) and is of length $2s + 1$ for some $s \in \bbN$.
    Since $G$ is connected, there exists a path from $i$ to $y$, one from $j$ to $y$, and one from $y$ to $x$.
    Let $d_{iy}$, $d_{jy}$, and $d_{yx}$ be the length of the shortest path from $i$ to $y$, from $j$ to $y$, and from $y$ to $x$, respectively.
    WLOG, suppose $d_{iy} \leq d_{jy}$.
    Now, set $K = N_1 + 2sN_2 + d_{jy} + d_{yx}$, and fix any $k \geq K$,
    since a length-$s$ cycle includes $y$, 
    there exists $q \leq d_{jy} + s$ such that there exist length-$q$ walks 
    $\mathbf{w}_{iy; q}$ and $\mathbf{w}_{jy; q}$
    from $i$ to $y$ and from $j$ to $y$, respectively,
    and $q + d_{yx} \equiv k \pmod 2$ (an additional cycle from $y$ back to $y$ via $\mathbf{c}$ can be added to ensure this).
    Now, fix any neighbor $z$ of $x$,
    for each $0 \leq r \leq N_2 - 1$,
    we can construct a pair of walks:
    it goes first from $i$ (resp., $j$) to $y$ via $\mathbf{w}_{iy; q}$ (resp., $\mathbf{w}_{jy; q}$), and then goes via $\mathbf{c}$ for $2r$ rounds ($2rs$ steps) back to $y$, and then goes to $x$ in $d_{yx}$ via the shortest path, and finally goes back and forth between $x$ and $z$ until the length of the walk reaches $k$.
    The degree of intersection between such a pair of walks is at least 
    $k - q \geq K - d_{jy} - s \geq (2s - 1)N_2 + d_{yx} + N_1 \geq N_1$,
    and the number of such pairs is at least the number of $r$ values, which is $N_2$,
    completing the proof.
\end{proof}

\subsection{Regarding the Number of Parameters}

\begin{proof}[Proof of Theorem~\ref{thm:number_params}]
    For \ours, the additional parameters include
    $W^{(k)}_{edge}$'s, $W^{(k)}_{hop}$'s, and $b^{(k)}_{hop}$'s, 
    where for each $k$,
    $W^{(k)}_{edge} \in \bbR^{2 d_H}$, $W^{(k)}_{hop} \in \bbR^{2 d_H}$, and $b^{(k)}_{hop} \in \bbR$.
    Therefore, the number of additional parameters is
    $k_{max} (4d_H + 1) = \Theta(k_{max} d_H)$.

    For FAGCN, the additional parameters include
    $W^{(k)}_{edge}$'s, where
    for each $k$, 
    $W^{(k)}_{edge} \in \bbR^{2 d_H}$.
    Therefore, the number of additional parameters is
    $k_{max} (2 d_H) = \Theta(k_{max}d_H)$.

    For GATv2, the additional parameters include
    $W^{(k)}$'s and $W^{(k)}_{edge}$'s, where for each $k$,
    $W^{(k)} \in \bbR^{d_H \times d_H}$ and $W^{(k)}_{edge} \in \bbR^{2 d_H}$.
    Therefore, the number of additional parameters is
    $k_{max} (d_H^2 + 2d_H) = \Theta(k_{max} d_H^2)$.
\end{proof}
Discussions on the complexity of other baselines can be found in Appendix~\ref{app:complexity}.

\subsection{A Relaxed yet Equivalent Variant of Definition~\ref{def:vul_to_over_smoothing}}\label{app:def_v2os_relax}
In Definition~\ref{def:vul_to_over_smoothing} in the main text, we have used the condition on the identity of the features of several nodes. Such a condition has been used for the ease and simplicity of presentation. However, we acknowledge that exactly equal node features are not common in practice, and thus, we provide here a relaxed yet equivalent variant of Definition~\ref{def:vul_to_over_smoothing}.

\begin{definition}[A relaxed yet equivalent variant of Definition~\ref{def:vul_to_over_smoothing}]
    Given $G = (V, E)$ and initial node features $X$ satisfying Assumption~\ref{aspt:basic},
    we say that $f_{att}$ is \bolita{vulnerable to over-smoothing} (V2OS), 
    if given any $\epsilon > 0$, there exists $\xi > 0$, such that
    $\forall \theta$, $\norm{\check{\alpha}^{(k'+1)}_{ij} - \check{\alpha}^{(k'+1)}_{i'j'}} \leq \epsilon$ and $\norm{\gamma^{(k')}_{i} - \gamma^{(k')}_{i'}} \leq \epsilon$, for any $(i, j), (i', j') \in E$ and $k' \geq 1$ with
    $\norm{H^{(k')}_{x} - H^{(k')}_{y}} \leq \alpha$, for each $\set{x, y} \in \binom{\set{i, j, i', j'}}{2}$;    
    we say that $f_{att}$ is \bolita{weakly resistant to over-smoothing} (WR2OS), 
    if given any $\epsilon, \xi > 0$ and any $(i, j), (i', j') \in E$ and $k' \geq 1$ with
    $\norm{H^{(k')}_{x} - H^{(k')}_{y}} \leq \xi$, for each $\set{x, y} \in \binom{\set{i, j, i', j'}}{2}$,
    there exists $\theta$, such that
    $\norm{\check{\alpha}^{(k'+1)}_{ij} - \check{\alpha}^{(k'+1)}_{i'j'}} > \epsilon$ or 
    $\norm{\gamma^{(k')}_{i} - \gamma^{(k')}_{i'}} > \epsilon$;
    and we say that $f_{att}$ is \bolita{strongly resistant to over-smoothing} (SR2OS), 
    if given any $\epsilon, \xi > 0$ and any $(i, j), (i', j') \in E$ and $k' \geq 1$ with
    $\norm{H^{(k')}_{x} - H^{(k')}_{y}} \leq \xi$, for each $\set{x, y} \in \binom{\set{i, j, i', j'}}{2}$,
    there exists $\theta$, such that
    $\norm{\check{\alpha}^{(k'+1)}_{ij} - \check{\alpha}^{(k'+1)}_{i'j'}} > \epsilon$ and
    $\norm{\gamma^{(k')}_{i} - \gamma^{(k')}_{i'}} > \epsilon$.
\end{definition}

\begin{remark}
    With the above variant of Definition~\ref{def:vul_to_over_smoothing}, Theorems~\ref{thm:vul_to_os} and \ref{thm:ours_resistant} still hold. We use Definition~\ref{def:vul_to_over_smoothing} in the main text for ease and simplicity of presentation.
\end{remark}

\begin{table*}[t]
\begin{center}
\caption{Node Classification Performance of GATs with and without Non-Linear Propagation} \label{tab:performance_gats}
    \resizebox{\textwidth}{!}{
    \setlength\tabcolsep{4pt}
    \renewcommand{\arraystretch}{1.0}
        \centering
        \begin{tabular}{c c c c c c c c c c c c c}
            \toprule
            
            \textbf{Dataset} & Chameleon & Squirrel & Actor & Texas & Cornell & Wisconsin & Computer & Photo & Wiki-CS & Pubmed & Citeseer & Cora 
            \vspace{0.2mm}
            \\

            \midrule
            
            \textbf{Homophily}  & 0.04 & 0.03 & 0.01 & 0.00 & 0.02 & 0.05 & 0.70 & 0.77 & 0.57 & 0.66 & 0.63 & 0.77 
            \vspace{0.2mm}
            \\
            \midrule
            \midrule
            \vspace{0.2mm}
            \textbf{GATv2}          & \textbf{69.06 $\pm$ 2.2} & \textbf{57.67 $\pm$ 2.4} & 30.27 $\pm$ 0.8 & 60.32 $\pm$ 7.0 & 58.35 $\pm$ 3.8 & \textbf{61.94 $\pm$ 4.7} 
                                     & 84.19 $\pm$ 1.2 & 89.87 $\pm$ 1.2 & \textbf{79.64 $\pm$ 0.5} & 79.12 $\pm$ 0.3 & 71.15 $\pm$ 1.2 & \textbf{83.88 $\pm$ 0.6} 
                                     \vspace{0.2mm}
                                     \\
            \textbf{GATv2 \textit{(Lin)}}  & 68.73 $\pm$ 2.0 & 57.54 $\pm$ 1.8 & \textbf{30.45 $\pm$ 1.0} & \textbf{60.81 $\pm$ 6.1} & \textbf{58.65 $\pm$ 4.8} & 60.20 $\pm$ 5.8 
                                     & \textbf{85.01 $\pm$ 0.8} & \textbf{90.02 $\pm$ 0.8} & 79.53 $\pm$ 0.5 & \textbf{79.19 $\pm$ 0.4} & \textbf{71.40 $\pm$ 1.0} & 83.87 $\pm$ 0.5 
                                     \vspace{0.2mm}
                         \\

            \midrule
            \textbf{GATv2$^R$}  & 70.88 $\pm$ 1.9 & \textbf{61.23 $\pm$ 1.5} & \textbf{33.73 $\pm$ 0.9} & 60.68 $\pm$ 6.6 & 57.32 $\pm$ 4.5  & \textbf{60.61 $\pm$ 5.1} 
                         & 81.73 $\pm$ 2.2 & 88.71 $\pm$ 1.7 & \textbf{79.75 $\pm$ 0.6} & 78.28 $\pm$ 0.4 & 71.00 $\pm$ 0.8  & \textbf{82.42 $\pm$ 0.6} 
                         \vspace{0.2mm}
                         \\
            \textbf{GATv2$^R$ \textit{(Lin)}}  & \textbf{70.92 $\pm$ 1.7} & 60.40 $\pm$ 1.6 & 33.67 $\pm$ 0.7 & \textbf{61.35 $\pm$ 7.2} & \textbf{58.92 $\pm$ 3.6}  & 56.08 $\pm$ 5.8 
                         & \textbf{83.13 $\pm$ 1.5} & \textbf{89.42 $\pm$ 1.1} & 79.68 $\pm$ 0.8 & \textbf{78.44 $\pm$ 0.4} & \textbf{71.36 $\pm$ 0.7}  & \textbf{82.42 $\pm$ 0.5} 
                         \vspace{0.2mm}
                         \\
            \midrule

            \textbf{GT}  & \textbf{69.34 $\pm$ 1.2} & \textbf{55.04 $\pm$ 1.9} & \textbf{36.29 $\pm$ 1.0} & 84.08 $\pm$ 5.6 & \textbf{80.00 $\pm$ 4.9} & \textbf{84.80 $\pm$ 4.3}
                         & 84.38 $\pm$ 1.3 & \textbf{91.28 $\pm$ 1.1} & \textbf{79.93 $\pm$ 0.5} & \textbf{79.04 $\pm$ 0.5} & 70.16 $\pm$ 0.8 & \textbf{82.09 $\pm$ 0.7} 
                         \vspace{0.2mm}
                         \\
            \textbf{GT \textit{(Lin)}}  & 66.75 $\pm$ 1.4 & 49.46 $\pm$ 2.0 & 36.24 $\pm$ 0.9 & \textbf{84.59 $\pm$ 5.4} & 79.73 $\pm$ 4.9 & 83.33 $\pm$ 5.3
                         & \textbf{85.29 $\pm$ 1.2} & 90.58 $\pm$ 1.7 & 79.88 $\pm$ 0.5 & 78.93 $\pm$ 0.5 & \textbf{70.35 $\pm$ 0.6} & 81.97 $\pm$ 0.6 
                         \\



            \bottomrule
        \end{tabular}
        }
    \begin{tablenotes}
      \item \hspace{1mm} \textbullet{ \textit{(Lin)} denotes the linear propagation version of the model.}
      
      \item \hspace{1mm} \textbullet{ In each column, \textbf{bold} indicates higher performance between the two versions of the GATs.}
    \end{tablenotes}
\end{center}
\vspace{3mm}
\end{table*}

\section{Rephrased Propagation}\label{app:rewrite_msg_passing}
\noindent
In this section, we explain how we rephrase the propagation formulae for GATv2 and FAGCN and obtain the formulae in Table~\ref{tab: T}.

\smallsection{GATv2.} 
Many prior studies on GNNs have dropped non-linearity for their theoretical analyses ~\cite{cgat, oversmoothing, nonlinearity-removal(CCNS), nonlinearity-removal(ODG), dagnn, gprgnn}. 
In defining cumulative attention $T^{(k)}$, we also remove GATv2's non-linearity \textit{in the propagation layers}, 
while the non-linearity in computing edge attention coefficients $a^{(k)}_{ij}$'s is kept.
The detailed rephrasing is 
\[
H^{(k)} = \sigma(\mathcal{A}^{(k)}H^{(k-1)}W^{(k)}) \to \mathcal{A}^{(k)}H^{(k-1)}W^{(k)}
\]
We define GATv2's $T^{(k)}$ from this rephrase, such that $T^{(k)} =  \prod\nolimits_{\ell = k}^1 \mathcal{A}^{(\ell)}$.
\footnote{$\Gamma^{(k)}$ is removed because it is the identity matrix for GATv2. Recall that $T^{(k)}$ intuitively expresses attention between all node pairs at layer $k$. Thus, the definition of cumulative attention $T^{(k)}$ after rephrase assumes linear propagation, such that the attention GATv2 expresses for $k$ hop neighbors is equal to $\prod\nolimits_{l = k}^1 \mathcal{A}^{(l)}$}

We claim that removing non-linearity in the propagation layers may not significantly affect the performance and expressive power of GATv2, 
especially in the context of node classification.
In fact, two theoretical works have argued that non-linear propagation is not always beneficial in the context of node classification:
\begin{itemize}
    \item \citet{gnn_bayesian_nonlinear} use the Contextual Stochastic Block Model (CSBM) to show that non-linear propagation and linear propagation return similar performance, except when the node features are extremely informative.
    \item \citet{oversmoothing_nonlinear_2} also use CSBM, and they show that adding non-linearity after propagation does not improve model performance.
\end{itemize}

One theoretical work has argued for the universality of linear GNNs.
\begin{itemize}
    \item \citet{linear-gnn-power} show that GNNs without non-linearity are universal w.r.t. one-dimensional prediction, if there are no multiple eigenvalues of graph Laplacian or missing frequency components in node features.
\end{itemize}

Some empirical evidence also exists in line with the theoretical findings:
\begin{itemize}
    \item \citet{dagnn} report that the hidden node features become smooth at a quicker rate for GCN (a non-linear propagation model) than for DAGNN (a linear propagation model).
    \item \citet{evaluatedeepGNN} argues with empirical evidence that stacked non-linearity may relate to performance degradation in deep GNNs.
    \item Moreover, many GNNs achieve their state-of-the-art performance in node classification without non-linearity in their propagation layers, such as APPNP, GPRGNN, DAGNN, FAGCN, ASGC~\cite{asgc}, JacobiConv~\cite{linear-gnn-power}, and SlenderGNN~\cite{slender}. 
\end{itemize}

This does not mean that using non-linear propagation is futile. 
However, we argue that removing non-linearity from GATv2's propagation layer may not significantly compromise its performance and expressive power, especially in the context of node classification. 
Finally, in Table~\ref{tab:performance_gats}, we empirically evidence our claim by comparing the performance of original GATs to the rephrased, linear propagating GATs.
We observe that GATs with linear propagation have competitive performance.

\smallsection{FAGCN.} 
The original propagation equations of FAGCN are as follows:
\begin{align*}    
    H^{(0)} &= \sigma(W^{(1)}X) \\
    Z^{(k)} &= \Gamma Z^{(0)} + \mathcal{A}^{(k)}Z^{(k-1)} \\
    Z^* &= W^{(2)}Z^{(k)} 
\end{align*}
\begin{table*}[t]
\begin{center}
\caption{The basic statistics of the benchmark datasets} \label{tab:data_stat}
    \resizebox{\textwidth}{!}{
    \renewcommand{\arraystretch}{1.0}
        \centering
        \begin{tabular}{*{13}{c}}
            \toprule            
            \textbf{Dataset} & Chameleon & Squirrel & Actor & Texas & Cornell & Wisconsin & Computer & Photo & Wiki-CS & Pubmed & Citeseer & Cora \\
            \midrule
            \textbf{\# Nodes} & 2,277 & 5,201 & 7,600 & 183 & 183 & 251 & 13,752 & 7,650 & 11,701 & 19,717 & 3,327 & 2,708 \\
            \textbf{\# Edges} & 31,371 & 198,353 & 26,659 & 279 & 277 & 450 & 245,861 & 119,081 & 215,603 & 44,324 & 4,552 & 5,278 \\
            \textbf{\# Features} & 2,325 & 2,089 & 932 & 1,703 & 1,703 & 1,703 & 767 & 745 & 300 & 500 & 3,703 & 1,433  \\
            \textbf{\# Labels} & 5 & 5 & 5 & 5 & 5 & 5 & 10 & 8 & 10 & 3 & 6 & 7     \\
            \textbf{Homophily}  & 0.04 & 0.03 & 0.01 & 0.00 & 0.02 & 0.05 & 0.70 & 0.77 & 0.57 & 0.66 & 0.63 & 0.77 \\
            \textbf{Split Type} & Fixed & Fixed & Fixed & Fixed & Fixed & Fixed & Random & Random & Fixed & Fixed & Fixed & Fixed    \\
            \textbf{Data Split (\%)} & 48/32/20 & 48/32/20 & 48/32/20 & 48/32/20 & 48/32/20 & 48/32/20 & 2.5/2.5/95 & 2.5/2.5/95 & 5.0/15/50 & 0.3/2.5/5.0 & 5.2/18/37 & 3.6/15/30 \\
            \bottomrule
        \end{tabular}
        }
\end{center}
\vspace{3mm}

\end{table*}

We rephrase the above equations by
\begin{align*}    
        &Z^{(k)} \\
    =~&\Gamma H^{(0)} + \mathcal{A}^{(k)}Z^{(k-1)} \\
    =~&\Gamma H^{(0)} + \mathcal{A}^{(k)}(\Gamma H^{(0)} + \mathcal{A}^{(k-1)}Z^{(k-2)}) \\
    =~&\Gamma H^{(0)} + \Gamma \mathcal{A}^{(k)} H^{(0)} + \mathcal{A}^{(k)} \mathcal{A}^{(k-1)}Z^{(k-2)}) \\
    =~&\Gamma H^{(0)} + \Gamma \mathcal{A}^{(k)} H^{(0)} + \mathcal{A}^{(k)} \mathcal{A}^{(k-1)}(\Gamma H^{(0)} + \mathcal{A}^{(k-2)}Z^{(k-3)})) \\
    =~& \cdots \\    
    =~& (\Gamma + \Gamma \mathcal{A}^{(k)} + \Gamma \mathcal{A}^{(k)}\mathcal{A}^{(k-\ell)} +\cdots+ \Gamma \prod\nolimits_{\ell=k}^1 \mathcal{A}^{(\ell)} )H^{(0)}.
\end{align*}

\section{Benchmark Datasets}\label{app:datasets}
\subsection{Dataset Description}
We use 12 node classification benchmark datasets in our experiments:

\begin{itemize}
    \item The \textit{chameleon} and \textit{squirrel} datasets are webpage networks of Wikipedia~\cite{geomgcn, wiki-cham-squir},
    where each node represents a webpage on Wikipedia, and two nodes are adjacent if mutual links exist between the two corresponding web pages.
    For each node, the node features are informative nouns on the corresponding webpage,
    and the node label represents the category of the average monthly traffic of the corresponding webpage.
    \item The \textit{actor} dataset is the actor-only induced subgraph of a film-director-actor-writer network obtained from Wikipedia webpages~\cite{geomgcn, film_original},
    where each node represents an actor, and two nodes are adjacent if the two corresponding actors appear on the same Wikipedia webpage.
    For each node, the node features are derived from the keywords on the Wikipedia webpage of the corresponding actor,
    and the node label is determined by the words on the webpage.
    \item The \textit{texas}, \textit{cornell}, and \textit{wisconsin} datasets are extracted from the WebKB dataset~\cite{geomgcn},
    where each node represents a webpage, and two nodes are adjacent if a hyperlink between the two corresponding webpages.
    For each node, the node features are the bag-of-words features of the corresponding webpage,
    and the node label is the category of the webpage.    
    \item The \textit{computer} and \textit{photo} datasets are Amazon co-purchase networks~\cite{amazon},
    where each node represents a product, and two nodes are adjacent if the two corresponding products are frequently purchased together. 
    For each node, the node features are the bag-of-words features of the customer reviews of the corresponding product,
    and the node label is the category of the product.
    \item The \textit{wiki-cs} dataset is a webpage network of Wikipedia~\cite{wikics}, 
    where each node represents a Wikipedia webpage related to computer science,
    and two nodes are adjacent if a hyperlink exists between the two corresponding webpages. 
    For each node, the node features are the GloVe word embeddings of the corresponding webpage, 
    and the node label represents the article category of the webpage.
    \item The \textit{pubmed}, \textit{citeseer}, and \textit{cora} datasets are citation networks ~\cite{planetoid},
    where each node represents a research article, and two nodes are adjacent if a citation exists between the two corresponding articles.
    For each node, the node features are the bag-of-words features of the corresponding article,
    and the node label is the category of the research domain of the article.    
\end{itemize}

Following~\citet{gprgnn}, we preprocess the directed graphs among the datasets to be undirected, which is commonly done for node classification.
For measuring the degree of homophily of each dataset, we use the homophily metric suggested by~\citet{link}. 
In Table~\ref{tab:data_stat}, we present the basic statistics of the benchmark datasets after preprocessing.

\subsection{Train/Val/Test Split}
Experiments are generally conducted under the most commonly used settings. Two different training schemes are used: sparse- and dense-labeled training. Sparse- and dense-labeled training, respectively, refer to having small and large proportions of train labels. In line with the prior works, we conducted experiments with sparse-labeled training for the homophilic graphs~\cite{gcn, gat, appnp, dagnn} and dense-labeled training for the heterophilic graphs~\cite{geomgcn, gprgnn, fagcn}. 

Overall, we used publicly available splits. The public splits of \textit{cora}, \textit{citeseer}, and \textit{pubmed} were given by ~\cite{planetoid}, and \textit{wiki-cs} by ~\cite{wikics}. Since \textit{computer} and \textit{photo} datasets do not have publicly available splits, we follow the methodology of prior works~\cite{gprgnn} and use a random (2.5\%, 2.5\%, 95\%) split for each trial. For heterophilic graphs, the public splits were provided by \citet{geomgcn}.
\section{Hyperparameters}\label{app:methods_params}

In this section, we detail the hyperparameter search space of the 14 GNN models used in the experiments.
Overall, for each baseline model, we try to maintain the experimental settings and hyperparameters search spaces used in the original paper.

\smallsection{Learning Rate and Weight Decay.}
For all the methods, we set the learning rate as $0.01$ for sparse-labeled training and $0.005$ for dense-labeled training. 
For models with separate decay weights for parameters in propagation and feature transformation layers, we use $\text{WD}_{prop}$ and $\text{WD}_{ft}$ to denote each. 

\smallsection{Hidden Dimension $d_H$.}
We generally maintain the same hidden dimension sizes for all models.
For GAT, GATv2, GAT-Res, and GT, we use 8 attention heads and set the hidden dimension as 8, so that the total hidden dimension = 64.
For FAGCN, we follow the original paper, setting the hidden dimension = 16 for homophilic graphs and setting the hidden dimension = 32 for heterophilic graphs.
For others, the hidden dimension is set to 64.

\smallsection{Number of Layers $k_{max}$.}
For APPNP and GPRGNN, the number of propagation layers is set as 10, as proposed in the original papers. 
For GCN, we use two layers.
Mixhop uses two dense linear layers, and we search for the best number of propagation layers per linear layer.
For \ours, we use a two-layer MLP to obtain $H^{(0)}$ for dense-labeled training, while a one-layer MLP is used for the sparse-labeled training. 
For others, we search their best-performing layer.

\smallsection{Dropout.}
For \ours, an additional dropout layer is optionally added before the last linear layer to obtain $Z^*$ from $Z^{(k_{max})}$, when doing so is helpful for the performance (specifically, in \textit{actor}, \textit{texas}, \textit{photo}, \textit{wiki-cs}, \textit{citeseer}, and \textit{cora} datasets).

\smallsection{Search Space.}
For each model, we choose the hyperparameters with which the model returns the best mean performance across 5 fixed random seeds.
For the search spaces with a combinatorial size of below 300, we use grid search. 
For the search spaces consisting of at least 300 different combinations, (including GATv2$^R$, FAGCN, GPRGNN, GCN-II, A-DGN, and \ours), 
we run 300 search trials using a Bayesian optimization algorithm.

Below, we list the hyperparameter search space for each model:

\begin{enumerate}
    \item \textbf{GCN}: 
    \begin{itemize}[leftmargin=-0.1mm]
        \item $\text{WD} \in \{1e-2, 5e-3, 1e-3, 5e-4, 1e-4\}$
        \item $\text{dropout} \in \{0.5, 0.6, 0.7, 0.8\}$
        \item $\text{L2} \in \{0.0, 5e-4\}$
    \end{itemize}   
    \item \textbf{GAT, GATv2, GT, DMP}:
    \begin{itemize}[leftmargin=-0.1mm]
        \item $\text{WD} \in \{1e-2, 5e-3, 1e-3, 5e-4, 1e-4\}$
        \item $\text{dropout} \in \{0.5, 0.6, 0.7, 0.8\}$
        \item $k_{max} \in \{2, 3\}$
        \item $\text{L2} \in \{0.0, 5e-4\}$
    \end{itemize}    
    \item \textbf{GATv2$^R$}: 
    \begin{itemize}[leftmargin=-0.1mm]
        \item $\text{WD}$ $\in$ \{1e-2, 5e-3, 1e-3, 5e-4, 1e-4\}
        \item $\text{dropout} \in \{0.5, 0.6, 0.7, 0.8\}$
        \item $k_{max} \in \{2, 4, 8, 16, 32\}$
        \item residual connection weight $\alpha \in \{0.1, 0.2, 0.3, 0.4, 0.5\}$
        \item L2 $\in \{0.0, 5e-4\}$
    \end{itemize}
    \item \textbf{FAGCN}: 
    \begin{itemize}[leftmargin=-0.1mm]
        \item $\text{WD}$ $\in \{1e-2, 5e-3, 1e-3, 5e-4, 1e-4\}$
        \item dropout $\in \{0.4, 0.5, 0.6, 0.7, 0.8\}$
        \item $k_{max}$ $\in \{1, 2, 3, 4, 5, 6, 7, 8\}$
        \item residual connection weight $\epsilon$ $\in$ $\{0.1, 0.2, ..., 0.9, 1.0\}$
        \item L2 $\in$ $\{0.0, 5e-4\}$
    \end{itemize}    
    \item \textbf{APPNP}: 
    \begin{itemize}[leftmargin=-0.1mm]
        \item $\text{WD}$ $\in$ $\{1e-2, 5e-3, 1e-3, 5e-4, 1e-4\}$
        \item dropout $\in$ $\{0.5, 0.6, 0.7, 0.8\}$
        \item L2 $\in$ $\{0.0, 5e-4\}$
        \item return probability $\alpha$ $\in$ $\{0.1, 0.3, 0.5, 0.9\}$
    \end{itemize}
    \item \textbf{GPRGNN}: 
    \begin{itemize}[leftmargin=-0.1mm]
        \item $\text{WD}$ $\in$ $\{1e-2, 5e-3, 1e-3, 5e-4, 1e-4\}$
        \item dropout $\in$ $\{0.5, 0.6, 0.7, 0.8\}$
        \item $k_{max}$ $\in$ $\{4, 8, 16, 32\}$
        \item L2 $\in$ $\{0.0, 5e-4\}$
        \item return probability $\alpha$ $\in$ $\{0.1, 0.3, 0.5, 0.9\}$    
    \end{itemize}
    \item \textbf{DAGNN}: 
    \begin{itemize}[leftmargin=-0.1mm]
        \item $\text{WD}$ $\in$ $\{0, 2e-2, 1e-2, 5e-3, 1e-3, 5e-4, 1e-4, 5e-5, 1e-5\}$
        \item dropout $\in$ $\{0.5, 0.6, 0.7, 0.8\}$
        \item $k_{max}$ $\in$ $\{5,10,20\}$
        \item L2 $\in$ $\{0.0, 5e-4\}$    
    \end{itemize}
    \item \textbf{MixHop}: 
    \begin{itemize}[leftmargin=-0.1mm]
        \item $\text{WD}$ $\in$ $\{1e-2, 5e-3, 1e-3, 5e-4, 1e-4\}$
        \item dropout $\in$ $\{0.5, 0.6, 0.7, 0.8\}$
        \item $k_{max}$ $\in$ $\{6,10\}$\footnote{$k_{max}$ of Mixhop is the number of fully-connected layers (2) $\times$ the number of propagation layers (3 or 5)}
        \item L2 $\in$ $\{0.0, 5e-4\}$
    \end{itemize}    
    \item \textbf{GCN-II}: 
    \begin{itemize}[leftmargin=-0.1mm]
        \item $\text{WD}_{ft}$ $\in$ $\{1e-3, 5e-4, 1e-4, 5e-5, 1e-5, 5e-6, 1e-6\}$
        \item $\text{WD}_{prop}$ $\in$ $\{1e-2, 5e-3, 1e-3, 5e-4, 1e-4\}$
        \item dropout $\in$ $\{0.5, 0.6, 0.7, 0.8\}$
        \item $k_{max}$ $\in$ $\{4,8,16,32,64\}$
        \item residual connection weight $\alpha$ $\in$ $\{0.1, 0.2, 0.3, 0.4, 0.5\}$
        \item weight decay $\lambda$ $\in$ $\{0.5, 1.0, 1.5\}$
        \item L2 $\in$ $\{0.0, 5e-4\}$
    \end{itemize}    
    \item \textbf{A-DGN}: 
    \begin{itemize}[leftmargin=-0.1mm]
        \item $\text{WD}_{ft}$ $\in$ $\{1e-3, 5e-4, 1e-4, 5e-5, 1e-5, 5e-6, 1e-6\}$
        \item $\text{WD}_{prop}$ $\in$ $\{1e-2, 5e-3, 1e-3, 5e-4, 1e-4\}$
        \item dropout $\in$ $\{0.0, 0.3, 0.5, 0.7\}$
        \item $k_{max}$ $\in$ $\{2,4,8,16,32\}$
        \item discretization step $\epsilon$ $\in$ $\{1, 0.1, 0.01, 0.001, 0.0001\}$
        \item diffusion strength $\lambda$ $\in$ $\{1, 0.1, 0.01, 0.001, 0.0001\}$
        \item L2 $\in$ $\{0.0, 5e-4\}$
    \end{itemize}    
    \item \textbf{\ours}: 
    \begin{itemize}[leftmargin=-0.1mm]
        \item $\text{WD}_{ft}$ $\in$ $\{4e-2, 2e-2, 1e-2, 5e-3, 1e-3, 5e-4, 1e-4\}$
        \item $\text{WD}_{prop}$ $\in$ $\{2e-2, 1e-2, 5e-3, 1e-3, 5e-4, 1e-4\}$
        \item dropout $\in$ $\{0.5, 0.6, 0.7, 0.8\}$
        \item $k_{max}$ $\in$ $\{4, 8, 16, 32\}$\footnote{\ours's performance sometimes marginally increase when $k_{max} = 64$, but it often saturates between $k_{max}$ of 16 and 32. For efficient search, we did not include $k_{max} = 64$.}
        \item weight decay $\lambda$ $\in$ $\{0.25, 0.5, 1.0\}$
    \end{itemize}    
\end{enumerate}

\section{Model Complexity Analysis}\label{app:complexity}
In Theorem~\ref{thm:number_params}, we have discussed the number of parameters for \ours, FAGCN, and GATv2.
Below, we analyze the model complexity for all the baselines.
Recall that we ignore the parameters used in computing the first hidden features ($H^{(0)}$) and those in the output layer,
since they are used in all GNN models.
That is, we only consider the number of \textit{additional parameters} other than those parameters.

We observe the number of additional parameters used in \ours is comparable to, or even smaller than, those in the other representative GNNs.

\smallsection{GCN and GCN-II.}
The additional parameters include
$W^{(k)}$'s, where for each $k$,
$W^{(k)} \in \bbR^{d_H \times d_H}$.
Therefore, the number of additional parameters is
$k_{max} d_H^2 = \Theta(k_{max} d_H^2)$.

\smallsection{A-DGN.}
The additional parameters include
$2W$'s, 
where each
$W \in \bbR^{d_H \times d_H}$.
Therefore, the number of additional parameters is
$2d_H^2 = \Theta(d_H^2)$.

\smallsection{GAT, GATv2, and GATv2$^R$.}
The additional parameters include
$W^{(k)}$'s and $W^{(k)}_{edge}$'s, 
where for each $k$,
$W^{(k)} \in \bbR^{d_H \times d_H}$ and
$W^{(k)}_{edge} \in \bbR^{2 d_H}$.
Therefore, the number of additional parameters is
$k_{max} d_H^2 + k_{max}(2 d_H) = \Theta(k_{max} d_H^2)$.

\smallsection{GT.}
The additional parameters include
$W^{(k)}_1$'s, $W^{(k)}_2$'s, $W^{(k)}_3$'s, and $W^{(k)}_4$'s, 
where for each $k$,
$W^{(k)}$ $\in \bbR^{d_H \times d_H}$.
Therefore, the number of additional parameters is
$k_{max}(4 d_H^2) = \Theta(k_{max} d_H^2)$.

\smallsection{DMP.}
The additional parameters include
$W^{(k)}$'s and $W^{(k)}_{edge}$'s, 
where for each $k$,
$W^{(k)} \in \bbR^{d_H \times d_H}$ and
$W^{(k)}_{edge} \in \bbR^{d_H \times d_H}$.
Therefore, the number of additional parameters is
$k_{max}(2 d_H^2) = \Theta(k_{max} d_H^2)$.

\smallsection{FAGCN.}
The additional parameters include
$W^{(k)}_{edge}$'s, 
where for each $k$, 
$W^{(k)}_{edge} \in \bbR^{2 d_H}$.
Therefore, the number of additional parameters is
$k_{max}(2 d_H) = \Theta(k_{max}d_{H})$.
    

\smallsection{APPNP.}
APPNP does not require additional parameters after obtaining $H^{(0)}$. 
Therefore, the number of additional parameters is $0$.

\smallsection{DAGNN.}
The additional parameters include a $W_{hop} \in \bbR^{d_C}$, 
where $d_C$ = the number of labels.
Therefore, the number of additional parameters is
$d_C = \Theta(d_C)$.

\smallsection{GPRGNN}
The number of additional parameters is $\Theta(k_{max})$ since at each layer, a constant number of parameters are used to compute the attention coefficients.

\smallsection{\ours.}
The additional parameters include
$W^{(k)}_{edge}$'s, $W^{(k)}_{hop}$'s, and $b^{(k)}_{hop}$'s, where for each $k$,
$W^{(k)}_{edge} \in \bbR^{2 d_H}$, $W^{(k)}_{hop} \in \bbR^{2 d_H}$, and $b^{(k)}_{hop} \in \bbR$.
Therefore, the number of additional parameters is
$k_{max} (4d_H + 1) = \Theta(k_{max} d_H)$.

We conclude again that the number of additional parameters used in \ours is comparable to, or even smaller than, those in the other attention-based GNNs that use nontrivial edge attention.
Most baseline models use $\Theta(k_{max} d_H^2)$ additional parameters,
while \ours and FAGCN use $\Theta(k_{max} d_H)$ additional parameters.

\begin{table*}[t]
\begin{center}
\caption{\ours Performance with Multi-Head Attention} \label{tab:multihead}
    \resizebox{\textwidth}{!}{
    \setlength\tabcolsep{2.5pt}
    \renewcommand{\arraystretch}{1.0}
        \centering
        \begin{tabular}{c c c c c c c c c c c c c}
            \toprule
            
            \textbf{Dataset} & Chameleon & Squirrel & Actor & Texas & Cornell & Wisconsin & Computer & Photo & Wiki-CS & Pubmed & Citeseer & Cora 
            \vspace{0.2mm}
            \\
            \midrule            
            \textbf{Homophily}  & 0.04 & 0.03 & 0.01 & 0.00 & 0.02 & 0.05 & 0.70 & 0.77 & 0.57 & 0.66 & 0.63 & 0.77 
            \vspace{0.2mm}
            \\
            \midrule
            \midrule
            
            \textbf{\ours (1-head)} & {71.58 $\pm$ 2.4} 
                            & \textbf{61.76 $\pm$ 2.4} 
                            & \textbf{36.57 $\pm$ 1.1} 
                            & \textbf{84.35 $\pm$ 5.2} 
                            & {81.24 $\pm$ 6.8}
                            & \textbf{84.80 $\pm$ 3.3}
                            & \textbf{86.69 $\pm$ 1.4}
                            & \textbf{92.50 $\pm$ 0.7}
                            & \textbf{79.96 $\pm$ 0.5}
                            & \textbf{80.65 $\pm$ 0.5}
                            & \textbf{73.20 $\pm$ 0.6}
                            & \textbf{84.02 $\pm$ 0.5}
                            \vspace{0.2mm}
                            \\

            \textbf{\ours (4-head)} & \textbf{72.15 $\pm$ 1.7} 
                            & {61.50 $\pm$ 2.0} 
                            & {36.00 $\pm$ 1.3} 
                            & {82.78 $\pm$ 5.7} 
                            & \textbf{81.30 $\pm$ 7.2}
                            & {82.98 $\pm$ 4.9}
                            & {85.67 $\pm$ 1.4}
                            & \textbf{92.50 $\pm$ 0.8}
                            & {79.40 $\pm$ 0.5}
                            & {80.53 $\pm$ 0.5}
                            & {72.01 $\pm$ 1.2}
                            & {83.93 $\pm$ 0.6}
                            \vspace{0.2mm}
                            \\

            \bottomrule
        \end{tabular}
        }
\end{center}
\vspace{3mm}
\end{table*}

\section{Implementation Details}\label{app:implementation-details}
Here, we provide details about how the GNN models are implemented.

\smallsection{GAT, GATv2, FAGCN.}
We use official implementations from PyTorch Geometric~\cite{fey2019fast}. The dropout layer is applied to 
edge attention coefficients. The same weight matrix $W^{(k)}$ is applied to obtain both the source node and destination node features $H^{(k)}_i, H^{(k)}_j$.

\smallsection{GATv2$^R$.}
Following GCN-II and APPNP, we added an initial residual connection to GATv2. After obtaining $H^{(0)}$ from the initial feature matrix $X$, $H^{(0)}$ is added to hidden features of the next layers by residual weight $\epsilon$. Formally, GATv2$^R$ layer is defined as $Z^{(k)} = \sigma(((1-\epsilon) \mathcal{A}^{(k)}H^{(k)} + \epsilon H^{(0)})W^{(k)}).$

\smallsection{GT.}
GT computes edge attention coefficients with a transformer-like function. We used official implementation from PyTorch Geometric. The dropout layer is applied to edge attention coefficients. We do not use the gated function $\beta$ or Layer Normalization~\cite{layernorm}.

\smallsection{DMP.}
DMP computes vector edge attention coefficients, instead of a scalar attention coefficient, for each node pair. Then, it uses $\operatorname{tanh}$ to bound the edge attention coefficients, without further normalization. Since no official implementation is made public, we use our own implementation. Specifically, we used the DMP-1-Sum variant, where the neighbor information is sum-aggregated.

\smallsection{MixHop.}
MixHop concatenates the hidden feature $H^{(k)}$'s of different hops, which can be seen as a (simplified) form of hop attention. Simply put, each feature transformation layer comes after the concatenation of node features from multiple propagation layers. We used the implementation of MixHop from~\citet{linkx}.

\smallsection{GPRGNN.}
PageRank initialization is used for the hop attention coefficients $\gamma_i^{(k)}$'s. As originally implemented, we do not apply optimizer weight decay for the hop attention $\gamma^{(k)}_i$'s.

\smallsection{GCN-II.}
We use official implementations from PyTorch Geometric. GCN-II version with the shared weight matrix $W^{(k)}$ for initial and smoothed node feature ($H^{(0)}$ and $H^{(k)}$) is chosen.

\smallsection{A-DGN.}
We use official implementations from PyTorch Geometric. We used $\operatorname{tanh}$ activation and GCN layer as the message passing function.

\begin{table}[t]
\centering
\caption{Node Classification Performance on the Filtered Datasets} \label{tab:new_heterophily}
    \setlength\tabcolsep{3pt}
    \scalebox{0.85}
    {
        \centering
        \begin{tabular}{c c c}
            \toprule
            \textbf{Dataset} & Chameleon(F) & Squirrel(F) 
            \\
            \midrule
            \textbf{Homophily}  & 0.04 & 0.04 \\ 
            \midrule
            \midrule
            \textbf{GCN} & 43.29 $\pm$ 3.8 & 41.30 $\pm$ 1.9   
                         \\
            \textbf{APPNP} & 40.45 $\pm$ 3.1 & 39.18 $\pm$ 1.9 
                         \\
            \midrule

            \textbf{GCN-II}  & 43.01 $\pm$ 3.9 & 42.96 $\pm$ 2.6 
                         \\
            \textbf{A-DGN}  & \bfseries\colorbox{green}{46.60 $\pm$ 3.8} & \bfseries\colorbox{yellow}{44.70 $\pm$ 2.1} 
                         \\
            \midrule
            \textbf{GAT} & 41.06 $\pm$ 3.6 & 39.34 $\pm$ 1.9 
                         \\
            \textbf{GATv2} & 42.26 $\pm$ 2.9 & 37.93 $\pm$ 2.3 
                         \\
            \textbf{GATv2$^R$} & 41.38 $\pm$ 3.1 &  39.57 $\pm$ 1.7 
                         \\
            \textbf{GT}  & 40.94 $\pm$ 3.2 & 39.28 $\pm$ 1.9 
                         \\
            \textbf{FAGCN}   & \bfseries\colorbox{yellow}{46.14 $\pm$ 4.4} & 43.25 $\pm$ 2.2 
                         \\
            \textbf{DMP}   & 41.45 $\pm$ 4.6 & 41.28 $\pm$ 2.2 
                         \\
            \midrule
            \textbf{MixHop}   & 41.25 $\pm$ 4.2 & 39.44 $\pm$ 2.2 
                         \\
            \textbf{GPRGNN}   & 40.60 $\pm$ 3.3 & 40.62 $\pm$ 2.1 
                         \\
            \textbf{DAGNN}  & 40.30 $\pm$ 3.3 & 37.13 $\pm$ 2.0 
                         \\
            \midrule
            \textbf{\ours} & {43.30 $\pm$ 3.6} 
                            & \bfseries\colorbox{green}{46.23 $\pm$ 2.2} 
                         \\

            \bottomrule
        \end{tabular}
    }
      
\vspace{3mm}
\end{table}

\begin{figure}[htb]
    \centering   
    \includegraphics[scale=0.35]{icml2023/fig/camera_ready_v/ST_and_performance_legend_230526.pdf}\\
    \begin{subfigure}[a]{\linewidth}
         \centering
         \includegraphics[scale=0.28]{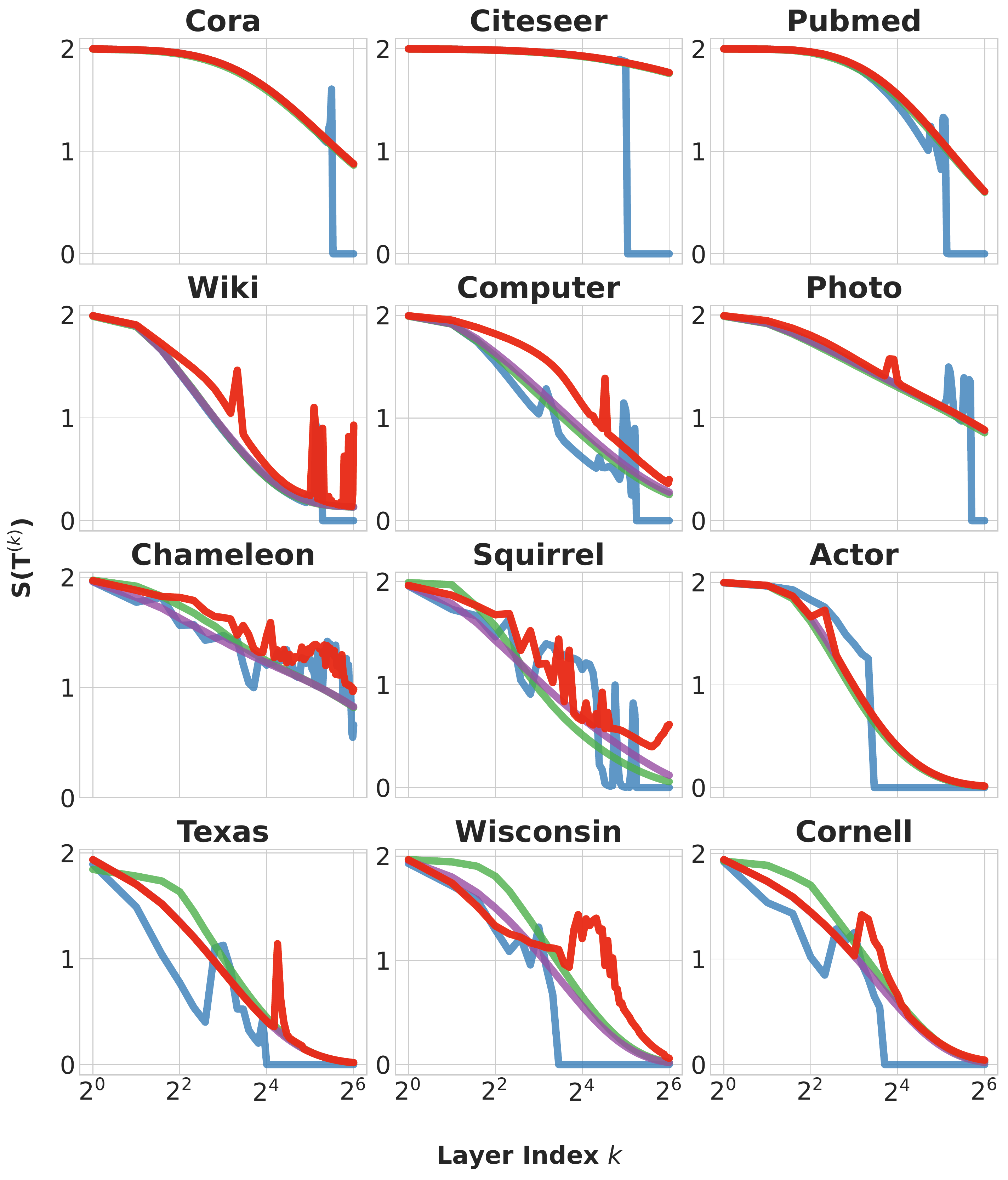}\\
         \vspace{-2mm}
    \end{subfigure}
    \caption{\label{fig:S(T)_all}
    \bolden{The smoothness score $S(T^{(k)})$ of the cumulative attention matrix $T^{(k)}$ for each $k$.}
    \textit{\ours has (1) less smoothed $T^{(k)}$ over deep layers and
    (2) often un-smoothes $T^{(k)}$.}
    } 
\end{figure}

\begin{figure}[htb] 
    \centering
    \hspace{10mm}
    \includegraphics[scale=0.35]{icml2023/fig/camera_ready_v/alpha_legend_230526.pdf}\\
    \begin{subfigure}[b]{\linewidth}
         \centering
         \includegraphics[scale=0.28]{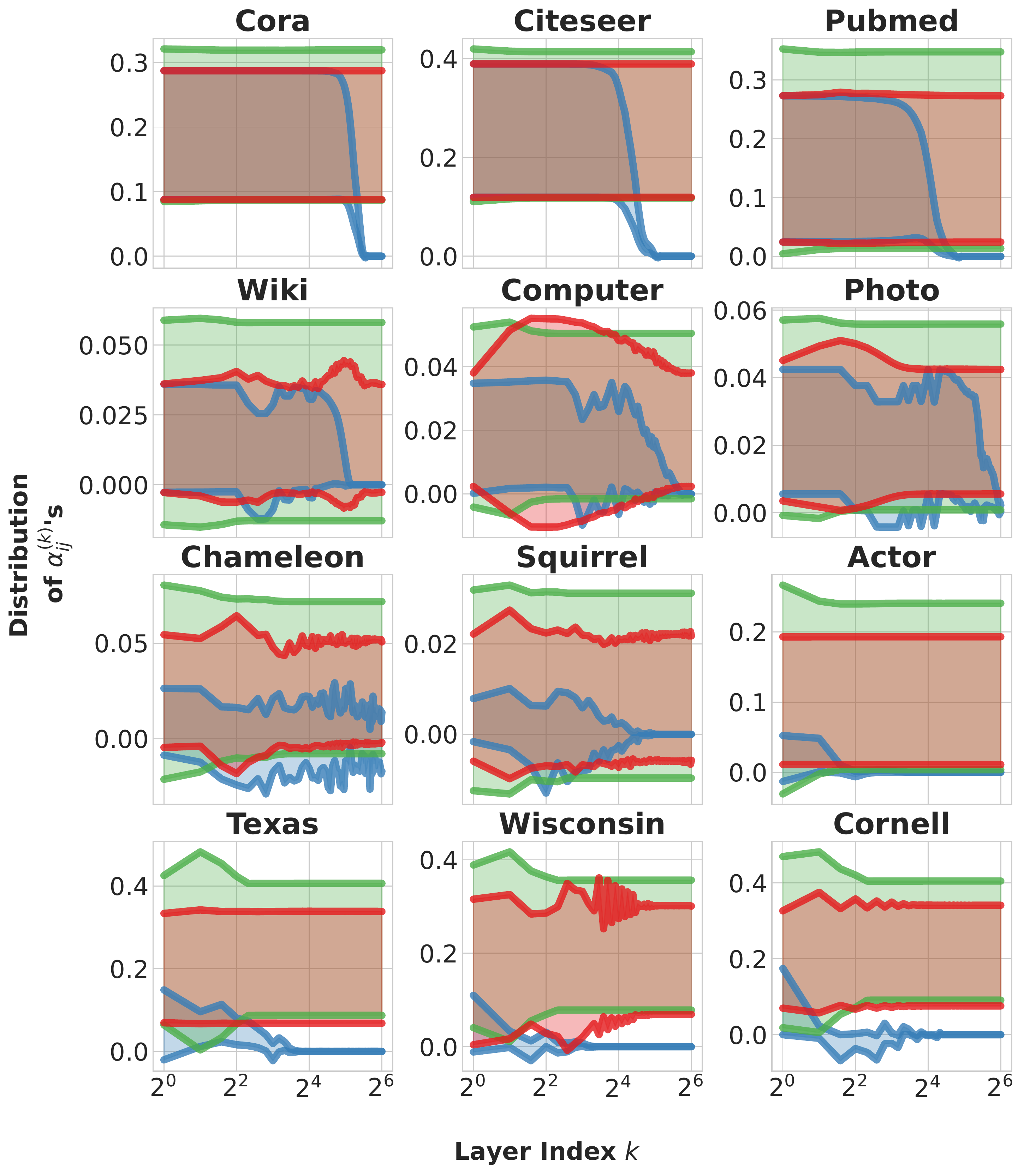}\\
         \vspace{-2mm}
         \caption{The distribution of $\alpha^{(k)}_{ij}$'s for each $k$. The shaded area represents the mean $\pm$ 1 SD.}
     \end{subfigure}
     \begin{subfigure}[b]{\linewidth}
         \centering
         \includegraphics[scale=0.28]{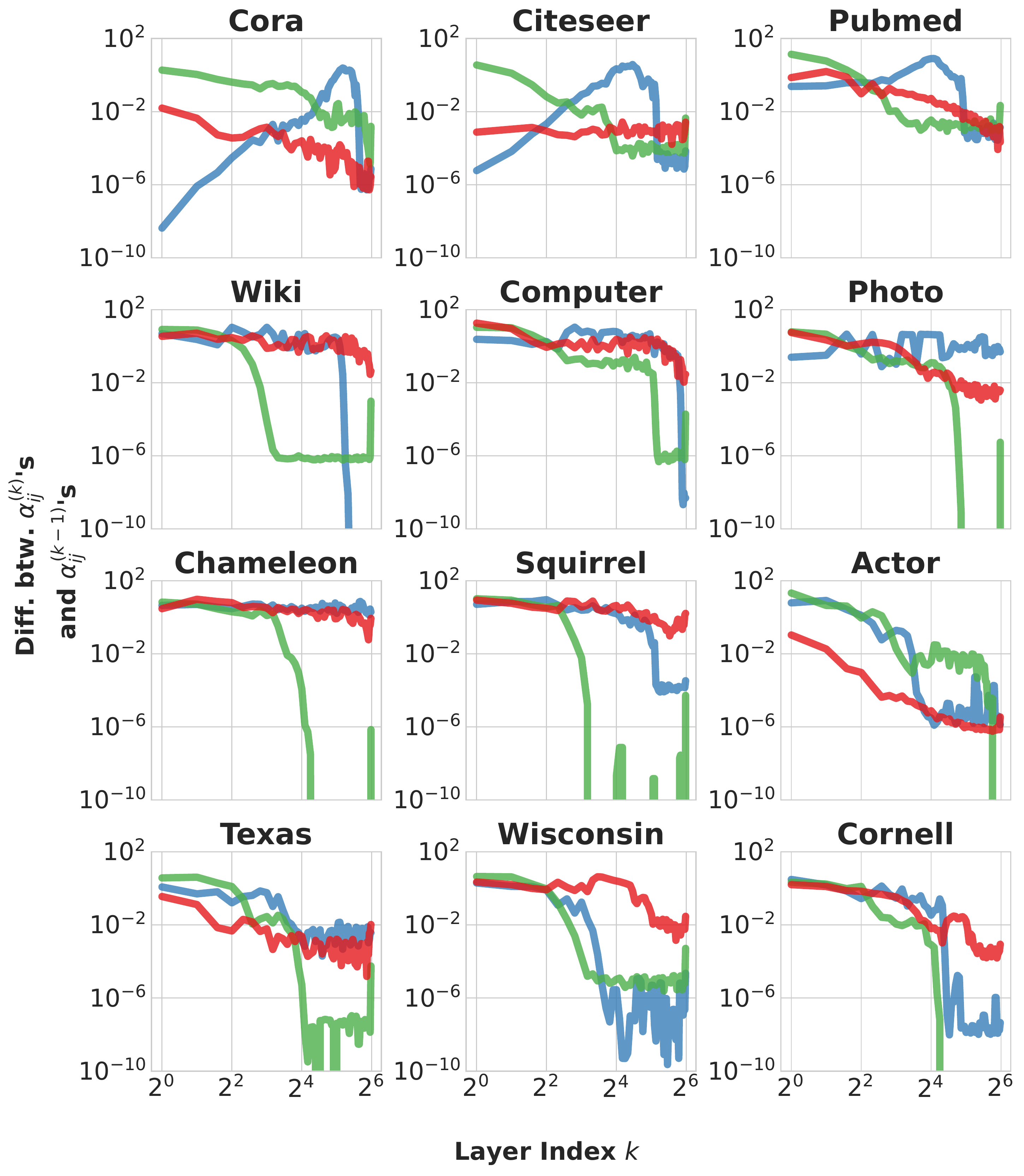}\\
         \vspace{-2mm}
         \caption{The differences between of $\alpha^{(k)}_{ij}$'s and $\alpha^{(k - 1)}_{ij}$'s for each $k$. 
         Formally, $\sqrt{\sum_{i, j} (\alpha^{(k)}_{ij} - \alpha^{(k - 1)}_{ij})^2}$ is reported for each $k$.}         
     \end{subfigure}    
    \caption{\label{fig:alpha_full}
    \bolden{Statistics w.r.t $\alpha^{(k)}_{ij}$'s for each $k$ with $k_{max}=64$ for all 12 datasets.}        
    }
\end{figure}

\begin{figure}[htb] 
    \centering
    \vspace{-13mm}
    \hspace{5mm}
    \includegraphics[scale=0.35]{icml2023/fig/camera_ready_v/gamma_legend_230123a.pdf}\\ 
    \begin{subfigure}[b]{\linewidth}
         \centering
         \includegraphics[scale=0.28]{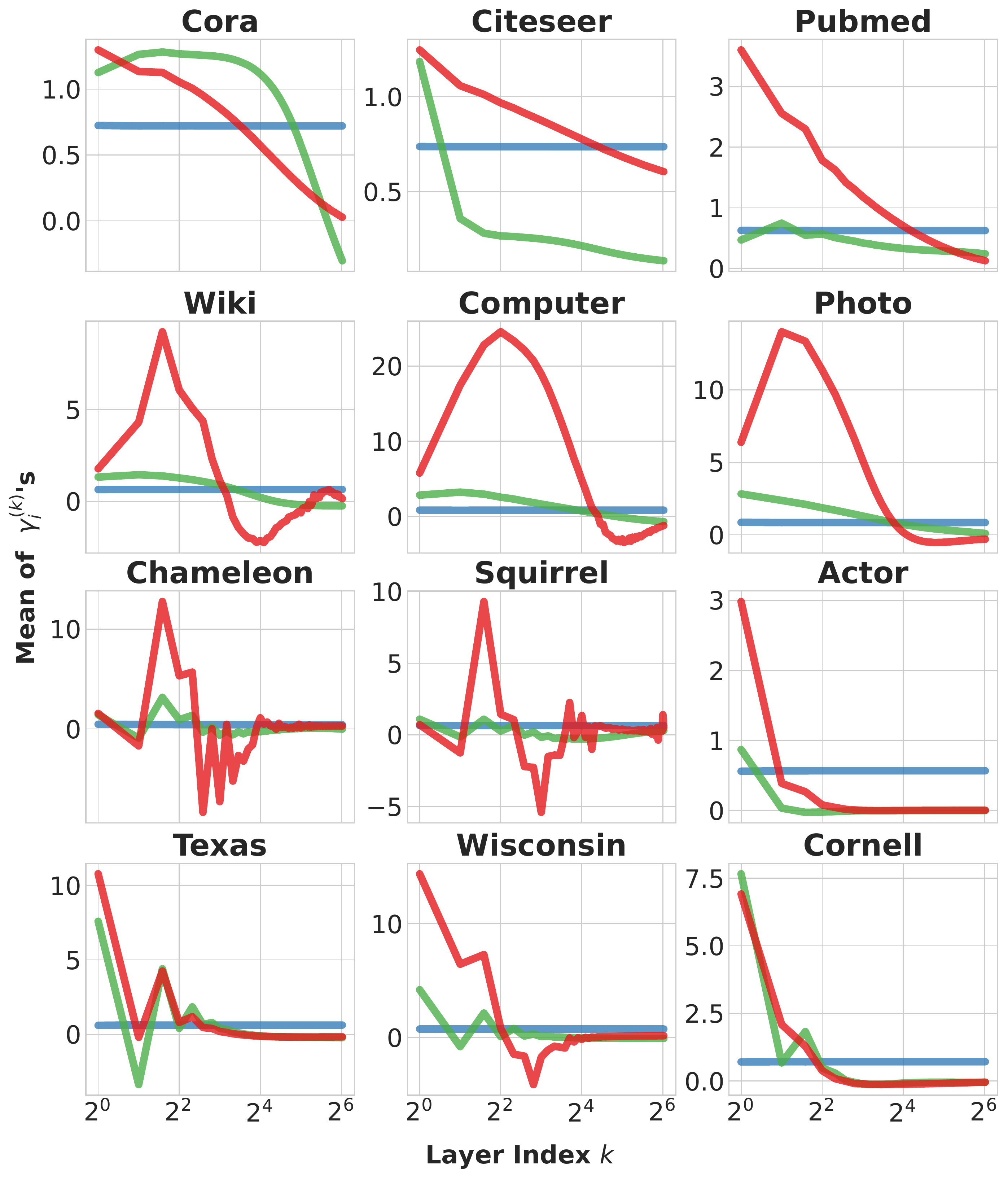}\\
         \vspace{-2mm}
         \caption{The mean of $\gamma^{(k)}_{i}$'s for each $k$.}         
     \end{subfigure}\\
     \begin{subfigure}[b]{\linewidth}
         \centering
         \includegraphics[scale=0.28]{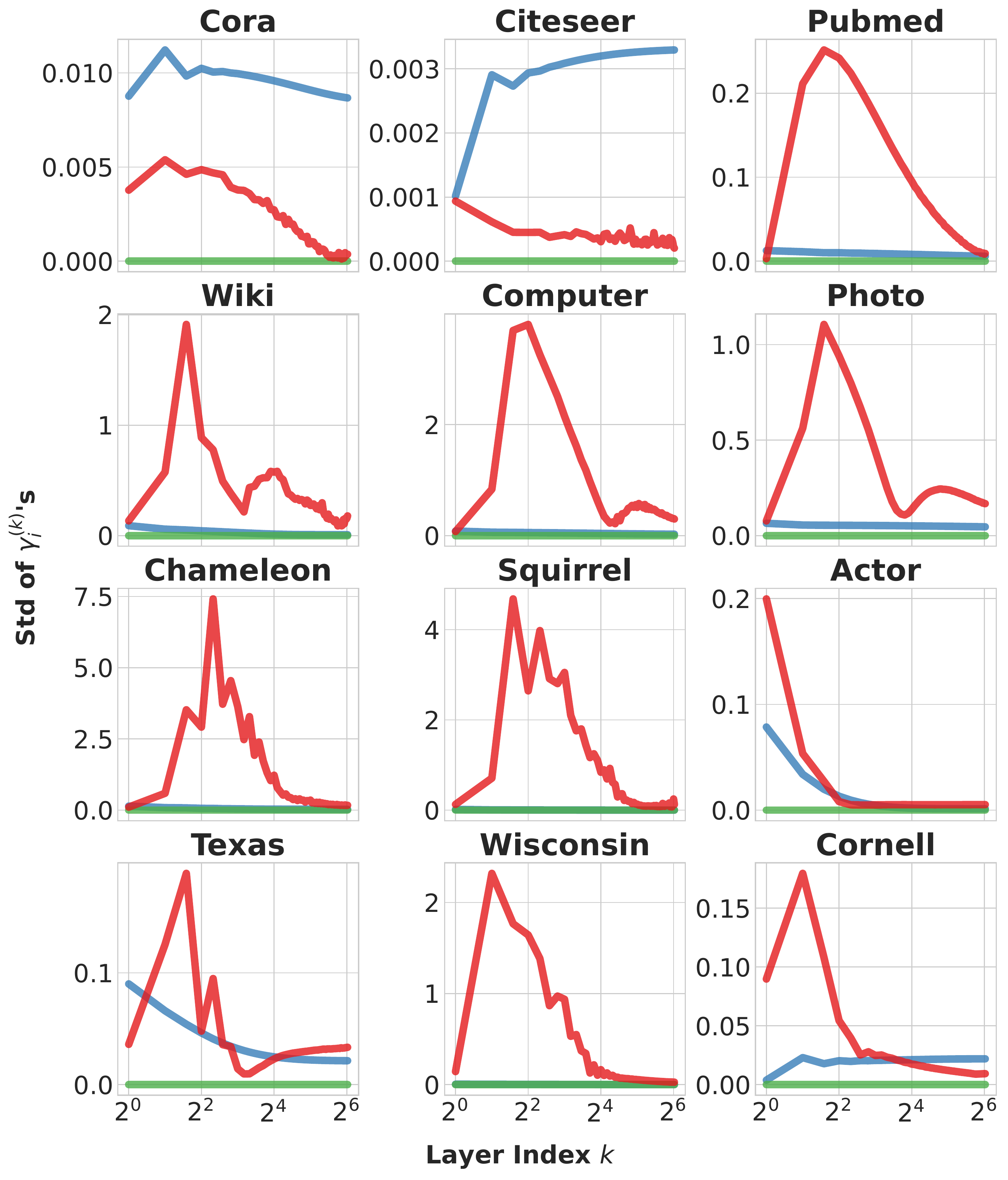}\\
         \vspace{-2mm}
         \caption{The SD of $\gamma^{(k)}_{i}$'s for each $k$.}         
     \end{subfigure}    
    \caption{\label{fig:gamma_full}    
    \bolden{Statistics w.r.t $\gamma^{(k)}_i$'s for each $k$ with $k_{max}=64$ for all 12 datasets.}
    }
\end{figure}

\section{Additional Experiments}\label{app:additional-exp}
\subsection{Generalization of \ours to Multi-Heads.}
We generalize \ours for multi-head attention by modifying the dimension of the attention functions. After obtaining $H^{(0)} \in \bbR^{n \times \eta \times d_H}$ with an MLP, where $\eta$ denotes the number of attention heads, dimension of attention weights are modified to $W^{(k)}_{edge} \in \bbR^{\eta \times 2 d_H}$ and $W^{(k)}_{hop} \in \bbR^{\eta \times 2 d_H}$. Subsequently, $\alpha^{(k)}_{ij}, \gamma^{(k)}_i \in \bbR^{\eta}$ and $H^{(k)}, Z^{(k)} \in \bbR^{n \times \eta \times d_H}$. At $k_{max}$, the multi-headed node features are concatenated to obtain the final node feature $Z^{(k_{max})} \in \bbR^{n \times (\eta \times d_H)}$.

We set the hidden dimension to be 16, with 4 attention heads. Consistent with the reports from \citet{gatv2}, we do not find multi-head attention to be always beneficial, while it occasionally improves performance. Refer to Table~\ref{tab:multihead} for details about performance with multi-head attention.

\subsection{Datasets Proposed in \citet{heterophily-eval}.}
Recently, \citet{heterophily-eval} argued that \textit{chameleon} and \textit{squirrel} datasets may have duplicate nodes and, consequently, train-test leakage problem. Therefore, they proposed to filter their duplicates for node classification. For a comprehensive evaluation, we conduct node classification on the filtered datasets (Table~\ref{tab:new_heterophily}). The same experimental settings are maintained. 

Consistent with the findings from \citet{heterophily-eval}, model performance drops significantly in the filtered datasets. Here, A-DGN performs the best, while AERO-GNN ranks second. It is still worth noting that AERO-GNN has the best average ranking among the attention-based GNNs.

\subsection{Empirical Evaluation of Cumulative Attention.}

As an extension of section ~\ref{exp: att-coef}, analyses of cumulative attention $T^{(k)}$ for \textit{all 12 datasets} are provided in Figure ~\ref{fig:S(T)_all}. Overall, the results are consistent with what is shown in Section~\ref{exp: att-coef}, and hence, we draw the same conclusions. AERO-GNN has consistently less smooth cumulative attention $T^{(k)}$ and often un-smoothes it.

\subsection{Empirical Analysis of Attention Coefficients.}
As an extension of section ~\ref{exp: att-coef}, analyses of attention coefficients for \textit{all 12 datasets} are provided in Figure ~\ref{fig:alpha_full} and ~\ref{fig:gamma_full}. Overall, the results are consistent with what is shown in section ~\ref{exp: att-coef}, and hence, we draw the same conclusions.

Besides, for Figure~\ref{fig:alpha} and~\ref{fig:alpha_full}, it is noteworthy that we reversed layer index $k$ (the x-axis) for GATv2$^R$ and FAGCN.  
Specifically, due to their initial residual connection, the two models have $T^{(k)} = \Gamma^{(k)} \prod\nolimits_{\ell = k_{max}}^{k_{max} - k + 1} \mathcal{A}^{(\ell)}$, 
while other models have $T^{(k)} =  \Gamma^{(k)} \prod\nolimits_{\ell = k}^1 \mathcal{A}^{(\ell)}$.
That is, for the two models, there is an \textit{inverse relationship} between the layer index $k$ and the corresponding hop that the attention coefficients learn to propagate
(recall that $T^{(k)}$ expresses attention between all node pairs within $k$ hops).

\begin{enumerate}
    \item \textbf{GATv2}: 
    \begin{itemize}[leftmargin=-0.1mm]
        \item \textbf{Fig.~\ref{fig:alpha_full} (a)}: Distributions of $\alpha^{(k)}_{ij}$'s remain stationary across layers $k$'s and similar across different graphs (\textit{not} edge-, graph-adaptive).
        \item \textbf{Fig.~\ref{fig:alpha_full} (b)}: Differences between $\alpha^{(k)}_{ij}$'s and $\alpha^{(k-1)}_{ij}$'s drop significantly to near 0 at large $k$ for some datasets (\textit{not} hop-adaptive).
    \end{itemize}   

    \item \textbf{FAGCN}: 
    \begin{itemize}[leftmargin=-0.1mm]
        \item \textbf{Fig.~\ref{fig:alpha_full} (a)}: Distributions of $\alpha^{(k)}_{ij}$'s (Fig. ~\ref{fig:alpha_full} (a)) shrink near 0 at large $k$ for all but \textit{chameleon} dataset (learning failure).
        \item \textbf{Fig.~\ref{fig:alpha_full} (b)}: Differences between $\alpha^{(k)}_{ij}$'s and $\alpha^{(k-1)}_{ij}$'s (Fig. ~\ref{fig:alpha_full} (b)) drop significantly to near 0 at large $k$ for some datasets (\textit{not} hop-adaptive).
    \end{itemize}   

    \item \textbf{GRPGNN}: 
    \begin{itemize}[leftmargin=-0.1mm]
        \item \textbf{Fig.~\ref{fig:gamma_full} (a)}: Means of $\gamma^{(k)}_{i}$'s are different for different hops and graphs (hop-, graph-adaptive).
        \item \textbf{Fig.~\ref{fig:gamma_full} (b)}: Standard deviations of $\gamma^{(k)}_{i}$'s are always 0 (\textit{not} node-adaptive).
    \end{itemize}   

    \item \textbf{DAGNN}: 
    \begin{itemize}[leftmargin=-0.1mm]
        \item \textbf{Fig.~\ref{fig:gamma_full} (a)}: Means of $\gamma^{(k)}_{i}$'s remain stationary across layers $k$ and graphs (\textit{not} hop-, graph-adaptive). 
        \item \textbf{Fig.~\ref{fig:gamma_full} (b)}: Standard deviations of $\gamma^{(k)}_{i}$'s are always very small (\textit{not} node-adaptive).
    \end{itemize}   

    \item \textbf{\ours}: 
    \begin{itemize}[leftmargin=-0.1mm]
        \item \textbf{Fig.~\ref{fig:alpha_full} (a)}: Distributions of $\alpha^{(k)}_{ij}$'s change across layers $k$ and are different across different graphs (edge-, graph-adaptive).
        \item \textbf{Fig.~\ref{fig:alpha_full} (b)}: Differences between $\alpha^{(k)}_{ij}$'s and $\alpha^{(k-1)}_{ij}$'s never drop significantly near 0. (hop-adaptive)
        \item \textbf{Fig.~\ref{fig:gamma_full} (a)}: Means of $\gamma^{(k)}_{i}$'s are different for different hops and graphs (hop-, graph-adaptive).
        \item \textbf{Fig.~\ref{fig:gamma_full} (b)}: Standard deviations of $\gamma^{(k)}_{i}$'s  are almost always high and have different distributions across different graphs (node-, graph-adaptive).
    \end{itemize}   
\end{enumerate}
Again, we claim that only the attention function of \ours can remain adaptive to edge/node, hop, and graph at deep layers.

\end{document}